\documentclass[10pt,twocolumn,letterpaper]{article}

\usepackage{cvpr}
\usepackage{times}
\usepackage{epsfig}
\usepackage{graphicx}
\usepackage{amsmath}
\usepackage{amssymb}
\usepackage{mathrsfs}
\usepackage{subfigure}
\usepackage[usenames,dvipsnames]{color}

\usepackage[pagebackref=true,breaklinks=true,letterpaper=true,colorlinks,bookmarks=false]{hyperref}
\definecolor{orange}{rgb}{0.1328125,0.54296875,0.1328125}
\hypersetup{
	colorlinks=true,
	linkcolor=red,
	filecolor=blue,      
	urlcolor=magenta,
	citecolor=cyan,
}

\cvprfinalcopy 

\def\cvprPaperID{6543} 

\ifcvprfinal\fi
\begin{document}

 
 \title{AutoTrack: Towards High-Performance Visual Tracking for UAV with Automatic Spatio-Temporal Regularization}
%

\author{Yiming Li${^\dagger}$, Changhong Fu$^{\dagger,*}$, Fangqiang Ding${^\dagger}$, Ziyuan Huang$^{\ddagger}$, and Geng Lu${^\S}$\\
${^\dagger}$Tongji University \ \ \ ${^\ddagger}$National University of Singapore \ \ \ ${^\S}$  Tsinghua University\\
	{\tt\small yimingli9702@gmail.com, changhongfu@tongji.edu.cn, lug@tsinghua.edu.cn}
}
\maketitle

\renewcommand{\thefootnote}{\fnsymbol{footnote}} 
\footnotetext[1]{Corresponding author} 

\thispagestyle{empty}

\begin{abstract}
Most existing trackers based on discriminative correlation filters (DCF) try to introduce predefined regularization term to improve the learning of target objects, e.g., by suppressing background learning or by restricting change rate of correlation filters. However, predefined parameters introduce much effort in tuning them and they still fail to adapt to new situations that the designer did not think of. In this work, a novel approach is proposed to online automatically and adaptively learn spatio-temporal regularization term. Spatially local response map variation is introduced as spatial regularization to make DCF focus on the learning of trust-worthy parts of the object, and global response map variation determines the updating rate of the filter. Extensive experiments on four UAV benchmarks have proven the superiority of our method compared to the state-of-the-art CPU- and GPU-based trackers, with a speed of $\sim$60 frames per second running on a single CPU. 

Our tracker is additionally proposed to be applied in UAV localization. Considerable tests in the indoor practical scenarios have proven the effectiveness and versatility of our localization method. The code is available at \url{https://github.com/vision4robotics/AutoTrack}.
\end{abstract}

\section{Introduction}
Visual object tracking is one of the fundamental tasks in the computer vision community, aiming to localize the object sequentially only with the information given in the first frame. Endowing unmanned aerial vehicle (UAV) with visual tracking capability brings many applications, \eg, aerial cinematography~\cite{Rogerio2019IROS}, person following~\cite{Rui2016CVPRW}, aircraft tracking~\cite{Fu2014ICRA}, and traffic patrolling~\cite{Karaduman2019JIRS}. 
\begin{figure}[t]
	\label{fig:scene}
	\includegraphics[width=\columnwidth]{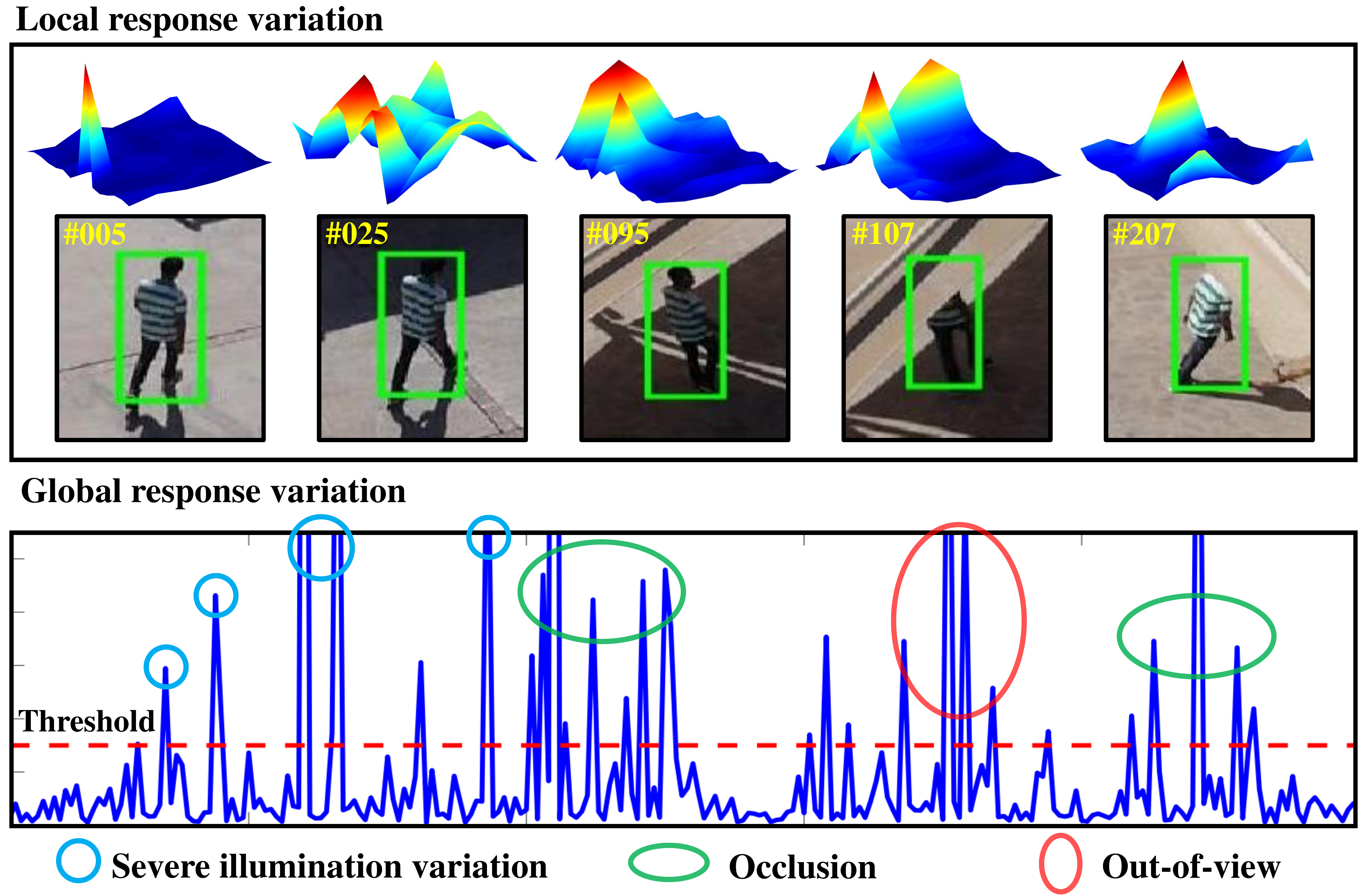}
	\caption{Central idea of our tracker. Spatially local and global response variations are exploited. Local variations indicate local credibility in the object bounding box. Severe illumination change in frame $25$ and $95$ as well as partial occlusion in frame $107$ and $207$ can lower the credibility of the appearance. AutoTrack is punished for learning these appearances so that local distractions can be avoided. In terms of global variations, large value can indicate wrong tracking result, where we stop the learning of correlation filters, while relatively large value should accelerate the learning of correlation filters so that adaptivity can be raised. }
\end{figure}

There are currently two main research interests in this area: discriminative correlation filter (DCF)-based methods~\cite{Bolme2010CVPR,Li2014ECCV,Henriques2015PAMI,Danelljan2015ICCV,Bertinetto2016CVPR,Mueller2017CVPR,Danelljan2017PAMI,Wang2017CVPR,Wang2018CVPR,Li2018CVPR} as well as  deep learning-based approaches~\cite{Bertinetto2016ECCV,Yun2017CVPR,Danelljan2016ECCV,Ma2015ICCV,Zhang2017CVPR,Danelljan2017CVPR}. In consideration of the limitation of power capacity and computational resources onboard UAVs, DCF framework  is selected because of its high efficiency originating from calculation in the Fourier domain. 

To improve DCF-based trackers, there are currently three directions: a) building more robust appearance model~\cite{Ma2015ICCV,Danelljan2016ECCV,Danelljan2017CVPR,fu2019correlation}, b) mitigating boundary effect or imposing restrictions in learning~\cite{Danelljan2015ICCV,Kiani2017ICCV,Danelljan2016ECCV,Li2018CVPR,Fu2019IROS}, and {c)} mitigating filter degradation~\cite{Danelljan2016CVPR,Wang2017CVPR,Li2018CVPR,LiICRA2020}. Robust appearance can indeed boost performance, yet it leads to burdensome calculations. Filter degradation, on the other hand, is not improving it fundamentally. Most trackers try to improve performance using option {b)} by introducing regularization terms. 

Recently, some attentions have been brought to using response maps generated in the detection phase to form the restrictions in learning~\cite{Huang2019ICCV}. The intuition behind it is that the response map contains crucial information regarding the resemblance of current object and the appearance model. However, \cite{Huang2019ICCV} only exploits what we call the spatially global response map variations, while ignoring local response variation indicating credibility at different locations in the image: drastic local variation means low credibility and vice versa. 

We fully exploit the local-global response variation to train our tracker with automatic spatio-temporal regularization, \ie, AutoTrack. While most parameters in regularization terms proposed by others are hyper-parameters that require large effort to tune, and would have a difficult time adjusting to new situations that the designers did not think of, we propose to learn some of the hyper-parameters automatically and adaptively. AutoTrack performs favorably against the state-of-the-art trackers, while running at $\sim$$60$ frames per second (fps) on a single CPU. 

Our main contributions are summarized as follows:   
\begin{itemize}
	\item We propose a novel spatio-temporal regularization term to simultaneously exploit local and global information hidden in response maps.
	\item We develop a novel DCF-based tracker which can automatically tune the hyper-parameters of spatio-temporal regularization term on the fly.
	\item We evaluate our tracker on 278 difficult UAV image sequences, and the evaluations have validated the state-of-the-art performance of our tracker compared to current CPU- and GPU-based trackers.
	\item We introduce a novel application of visual object tracking in UAV localization and prove its effectiveness as well as generality in the practical scenarios.
\end{itemize}
\section{Related Works}
\textbf{Tracking by detection:} tracking-by-detection framework, which regards the tracking as a classification problem, is widely adopted in UAV~\cite{Bolme2010CVPR,Li2014ECCV,Henriques2015PAMI,Zhang2013PR,Hare2016PAMI}. Among them, DCF has exhibited good performance with exceptional efficiency. The speed of traditional DCF-based trackers~\cite{Henriques2015PAMI,Bolme2010CVPR,Henriques2012ECCV} is around hundreds of fps on a single CPU, far exceeding the real-time requirement of UAV ($30$ fps). Yet they are primarily subjected to the following issues.

a) Boundary effect: the circulant samples suffer from periodical splicing at the boundary, reducing filters' discriminative power. Several works can mitigate boundary effect~\cite{Kiani2017ICCV,Danelljan2015ICCV,Li2018CVPR,LukezicCVPR2017}, but they used a constant spatial penalization which cannot adapt to various changes in different objects. K. Dai~\etal optimized the spatial regularization in the temporal domain~\cite{Dai2019CVPR}. Different to~\cite{Dai2019CVPR}, we exploit the inherent information in DCF framework, so our method is more generic. Also, we have achieved better performance in the aerial scenarios in terms of speed and precision. 

b) Filter degradation: the appearance model updated via a linear interpolation method cannot adapt to ubiquitous appearance change, leading to filter degradation. Some attempts are made to tackle the issue, \eg, training set management~\cite{Danelljan2017CVPR,Danelljan2016CVPR,LifanICRA2020}, temporal restriction~\cite{Li2018CVPR,LiICRA2020}, tracking confidence verification~\cite{Fu2019IROS,Wang2017CVPR} and over-fitting alleviation~\cite{Sun2019CVPR}. Amongst them the temporal regularization is an effective and efficient way. Yet the non-adaptive regularization is prone to tracking drift once the filter is corrupted. 

\textbf{Tracking by deep learning:}
recently, deep learning-based tracking has caught wide attention due to its robustness, \eg, deep feature representation~\cite{Ma2015ICCV,Danelljan2015ICCVWorkshop,Danelljan2016ECCV,Danelljan2017CVPR}, reinforcement learning~\cite{Yun2017CVPR}, residual learning~\cite{Song2017ICCV} and adversarial learning~\cite{songcvpr18VITAL}. However, for mobile robots, the above trackers cannot meet the requirement of real-time perception even with a high-end GPU. Currently, the state-of-the-art deep trackers~\cite{Li2018CVPR,LiBO2019CVPR,Danelljan2019ATOM,Zhang2019ICCV,Bhat2019ICCV,WangQiang2018CVPR} are mostly built on siamese neural network~\cite{Bertinetto2016ECCV}. The pre-trained siamese trackers just need to traverse in a feed-forward way to get a similarity score for object localization, facilitating real-time implementation on GPU. However, on a mobile device solely with CPU, the speed of siamese-based trackers cannot satisfy the real-time needs. C. Huang \etal proposed a CPU-friendly deep tracker~\cite{Huang2017ICCV} by training an agent working in a cascaded manner. It can run at near real-time speed by reducing calculation on easy frames. In summary, deep trackers can hardly meet real-time demands  on CPU. 

\textbf{Vision-based localization:}
vision-based localization is crucial for UAV especially in GPS-denied environments. A. Breitenmoser~\etal developed a monocular 6D pose estimation system based on passive markers in the visible spectrum~\cite{Breitenmoser2011IROS}. However, it performs worse in low-light environments. M. Faessler~\etal presented a monocular localization system based on infrared LEDs to raise robustness in cluttered environments~\cite{Faessler2014ICRA}. Its generality, however, is limited since the system can only work in the infrared spectrum. Built on~\cite{Faessler2014ICRA}, we develop a localization system based on visual tracking. In light of robustness and generality of our tracker in various scenarios like illumination variation, occlusion and deformation, our localization system is more versatile compared to the infrared LED-based one~\cite{Faessler2014ICRA}.

\section{Revisit STRCF}
In this section, our baseline STRCF~\cite{Li2018CVPR} is revisited. The optimal filter $\mathbf{H}_t$ in frame $t$ is learned by minimizing the following objective function:
	\begin{equation}\label{eq:strcf}
	\footnotesize
	\begin{aligned}
	\mathcal{E}(\mathbf{H_t})&=\frac{1}{2}\lVert\mathbf{y}-\sum_{k=1}^{K}\mathbf{x}_t^k \circledast \mathbf{h}_t^k\Vert_2^2+\frac{1}{2}\sum_{k=1}^{K}\Vert \mathbf{u}\odot \mathbf{h}^k_t\Vert_2^2\\
	&+\frac{\theta}{2}\sum_{k=1}^{K}\Vert\mathbf{h}_t^k-\mathbf{h}_{t-1}^k\Vert_2^2 
	\end{aligned} \ ,
	\end{equation}

\noindent where $\mathbf{x}^k_t\in \mathbb{R}^{T\times 1}(k=1,2,3,...,K)$ is the extracted feature with length $T$ in frame $t$, and $K$ denotes number of channel, $\mathbf{y}\in\mathbb{R}^{T\times 1}$ is the desired Gaussian-shaped response. $\mathbf{h}_t^k$, $\mathbf{h}_{t-1}^k\in\mathbb{R}^{T\times 1}$ respectively denote the filter of the $k$-th channel trained in the $t$-th and ($t$$-$$1$)-th frame, $\circledast$ indicates the convolution operator. Noted that $\mathbf{H}_t=[\mathbf{h}^1_t,\mathbf{h}^2_t,\mathbf{h}^3_t,...,\mathbf{h}^K_t]$. As for regularization, the spatial regularization parameter $\mathbf{u}\in\mathbb{R}^{T\times 1}$ is bowl-shaped and borrowed from SRDCF~\cite{Danelljan2015ICCV} for decreasing boundary effect, and temporal regularization, \ie, the third term in Eq.~\ref{eq:strcf}, is firstly proposed to restrict filter's variation by penalizing the difference between the current and previous filters.

Although STRCF~\cite{Li2018CVPR} has achieved competent performance, it does have two limitations: a) the fixed spatial regularization failing to address appearance variation in the unforeseeable aerial tracking scenarios, b) the unchanged temporal penalty strength $\theta$ (set as 15 in~\cite{Li2018CVPR}) which is not general in all kinds of situations. 

\section{Automatic Spatio-Temporal Regularization}
In this work, both local and global response variations are fully utilized to achieve simultaneous spatial and temporal regularizations, as well as automatic and adaptive hyper-parameter optimization. 
\subsection{Response Variation}
First of all, we define local response variation vector $\mathbf{\Pi}=[|\Pi^1|,|\Pi^2|,..., |\Pi^T|]$, as can be seen in Fig.~\ref{fig:scene} for its 2D visualization in the object bounding box, in preparation for spatial regularization. Its $i$-th element $|\Pi^i|$ is defined as:
\begin{equation}
	\small
	{\Pi^i}=\frac{\mathcal{R}_t[\psi_{\Delta}]^i-{\mathcal{R}}^i_{t-1}}{\mathcal{R}^i_{t-1}} \ ,
\end{equation} 
\noindent where $[\psi_{\Delta}]$ is the shift operator to make two peaks in two response maps ${\mathcal{R}}_{t}$ and ${\mathcal{R}}_{t-1}$ coincide with each other, in order for removing the motion influence~\cite{Huang2019ICCV}. $\mathcal{R}^i$ denotes the $i$-th element in response map $\mathcal{R}$.

\textbf{Automatic spatial regularization:} local response variation reveals the credibility of every pixel in the search area of the current frame. Therefore, filters located where the pixel credibility is low should be restricted in learning. We achieve this by introducing local variation $\mathbf{\Pi}$ to the spatial regularization parameter $\tilde{\mathbf{u}}$:
\begin{equation}\label{eq:spatial}
	\small
	\tilde{\mathbf{u}}=\mathbf{P}^\top\delta \log(\mathbf{\Pi}+1) + \mathbf{u} \ ,
\end{equation}	
\noindent where $\mathbf{P}^\top\in \mathbb{R}^{T\times T}$ is used to crop the central part of the filter where the object is located. $\delta$ is a constant to adjust the weight of local response variations, and $\mathbf{u}$ is inherited from STRCF~\cite{Li2018CVPR} to mitigate boundary effects. Through Eq.~\ref{eq:spatial}, filters located at pixels with dramatic response variation will be partially refrained from learning the new appearance because of the spatial punishment.

\textbf{Automatic temporal regularization:} in STRCF~\cite{Li2018CVPR}, the change rate of filters between two frames is punished in the loss by a fixed parameter $\theta$. AutoTrack tries to adaptively and automatically determine the value of this hyper-parameter by jointly optimization of its value and the filter. So we define a reference $\tilde{\theta}$ in preparation for the objective function with regard to the global response:
\begin{equation}\label{eq:temporal}
\small
	\tilde{\theta}=
	\begin{aligned}
	\frac{\zeta}{1+\log(\nu\Vert\mathbf{\Pi}\Vert_2+1)} , \ \ \ \Vert\mathbf{\Pi}\Vert_2\le\phi 
	\end{aligned} \ ,
\end{equation} 
\noindent where $\zeta$ and $\nu$ denote hyper parameters. When the global variation is higher than the threshold $\phi$, it means that there are aberrances in response maps~\cite{Huang2019ICCV}, so correlation filter ceases to learn. If it is lower than the threshold, the more dramatic the response map varies, the smaller the reference value will be, so that the restriction on temporal change of the correlation filters can be loosened and it can learn more rapidly in situations like large appearance variations.


\noindent\textbf{Remark 1:} Note that what we defined here is the reference value rather than the hyper-parameter itself. For the hyper-parameter of the temporal regularization, we use joint optimization to online estimate the value of it, so that the restriction can be online adaptively adjusted according to the response map variations. When appearance changes drastically, correlation filter learns more rapidly and vice versa. 

\subsection{Objective Optimization }	
Our objective function for joint optimization of filter as well as temporal regularization term can be written as:
	\begin{equation}
	\label{eq:objective}
	\small
	\begin{aligned}
	\mathcal{E}(\mathbf{H}_t,\theta_t)&=\frac{1}{2}\Vert \mathbf{y}-\sum_{k=1}^{K}\mathbf{x}_t^k\circledast\mathbf{h}^k_t\Vert_2^2+\frac{1}{2}\sum_{k=1}^{K}\Vert \tilde{\mathbf{u}}\odot \mathbf{h}_t^k\Vert^2_2\\
	&+\frac{\theta_t}{2}\sum_{k=1}^{K}\Vert \mathbf{h}_t^k-\mathbf{h}_{t-1}^k\Vert_2^2+\frac{1}{2}\Vert\theta_t-\tilde{\theta}\Vert_2^2
	\end{aligned}  \ ,
	\end{equation}
where $\tilde\theta$ and $\theta_t$  respectively denote the reference and optimized temporal regularization parameter, and $\tilde{\mathbf{u}}$ represents the automatic spatial regularization calculated via Eq.~\ref{eq:spatial}.

For optimization, we introduce an auxiliary variable $\mathbf{\widehat{\mathbf{g}}}_t$ by ordering $\widehat{\mathbf{g}}_t=\sqrt{T}\mathbf{F}{\mathbf{h}}_t (\mathbf{\widehat{G}}=[\mathbf{\hat{g}}^1_t,\mathbf{\hat{g}}^2_t,\mathbf{\hat{g}}^3_t,...,\mathbf{\hat{g}}^K_t])$ where $\mathbf{F}\in\mathbb{C}^{T\times T}$ denotes the orthonormal matrix and the symbol $\hat{  }$  denotes the discrete Fourier transform (DFT) of a signal. Then Eq.~\ref{eq:objective} is converted into the frequency domain: 
	\begin{equation}\label{eq:frequency}
	\small
	\begin{aligned}
	\mathcal{E}(\mathbf{H}_t,\theta_t,\widehat{\mathbf{G}}_t)&=\frac{1}{2}\Vert \mathbf{y}-\sum_{k=1}^{K}\widehat{\mathbf{x}}_t^k\odot\widehat{\mathbf{g}}^k_t\Vert_2^2+\frac{1}{2}\sum_{k=1}^{K}\Vert \tilde{\mathbf{u}}\odot \mathbf{h}_t^k\Vert^2_2\\
	&+\frac{\theta_t}{2}\sum_{k=1}^{K}\Vert \widehat{\mathbf{g}}_t^k-\widehat{\mathbf{g}}_{t-1}^k\Vert_2^2+\frac{1}{2}\Vert\theta_t-\tilde{\theta}\Vert_2^2 
	\end{aligned}		 \ .
	\end{equation}
	
By minimizing Eq.~\ref{eq:frequency}, an optimal solution can be obtained through alternating direction method of multipliers (ADMM)~\cite{Stephen2011FTML}. The Augmented Lagrangian form of equation Eq.~\ref{eq:frequency} can be formulated as:
	\begin{equation}\label{eq:aug}
	\small
	\begin{aligned}
	\mathcal{L}_t(\mathbf{H}_t,\theta_t,\widehat{\mathbf{G}}_t,\widehat{\mathbf{M}}_t)&=\mathcal{E}(\mathbf{H}_t,\theta_t,\widehat{\mathbf{G}}_t)+\frac{\gamma}{2}\sum_{k=1}^{K}\Vert\widehat{\mathbf{g}}_t^k-\sqrt{T}\mathbf{F}\mathbf{h}_t^k\Vert_2^2\\
	&+\sum_{k=1}^{K}(\widehat{\mathbf{g}}_t^k-\sqrt{T}\mathbf{F}\mathbf{h}_t^k)^\top\widehat{\mathbf{m}}_t^k
	\end{aligned} \ ,
	\end{equation}
where $\widehat{\mathbf{M}}_t=[\widehat{\mathbf{m}}_1,\widehat{\mathbf{m}}_2,...,\widehat{\mathbf{m}}_K]\in\mathbb{R}^{T\times K}$
is the Fourier transform of the Lagrange multiplier and $\gamma$ denotes the step size regularization parameter. By assigning $\mathbf{v}_t^k=\frac{\mathbf{m}_t^k}{\gamma}(\mathbf{V}_t^k=[\mathbf{v}_t^1,\mathbf{v}_t^2,...,\mathbf{v}_t^K
])$, Eq.~\ref{eq:aug} can be reformulated as:
	\begin{equation}\label{eq:8}
	\small
	\begin{aligned}
	\mathcal{L}_t(\mathbf{H}_t,\theta_t,\widehat{\mathbf{G}}_t,\widehat{\mathbf{V}}_t)&=\mathcal{E}(\mathbf{H}_t,\theta_t,\widehat{\mathbf{G}}_t)\\
	&+\frac{\gamma}{2}\sum_{k=1}^{K}\Vert\widehat{\mathbf{g}}_t^k-\sqrt{T}\mathbf{F}\mathbf{h}_t^k+\widehat{\mathbf{v}}_t^k\Vert_2^2 
	\end{aligned}  \ .
	\end{equation}

Then we solve the following subproblems by ADMM.

\textbf{Subproblem $\widehat{\mathbf{G}}$:} given $\mathbf{H}_t,\theta_t,\widehat{\mathbf{V}}_t$,  the optimal	$\widehat{\mathbf{G}}^*$ is: 
	\begin{equation}\label{eq:g}
	\small
	\begin{aligned}
	\widehat{\mathbf{G}}^*&=\mathop{arg \ min}\limits_{\widehat{\mathbf{G}}}\{\frac{1}{2}\Vert \widehat{\mathbf{y}}-\sum_{k=1}^{K}\widehat{\mathbf{x}}_t^k\odot\widehat{\mathbf{g}}_t^k\Vert_2^2\\
	&+\frac{\theta_t}{2}\sum_{k=1}^{K}\Vert\widehat{\mathbf{g}}_t^k-\widehat{\mathbf{g}}_{t-1}^k\Vert_2^2
	+\frac{\gamma}{2}\sum_{k=1}^{K}\Vert\widehat{\mathbf{g}}_t^k-\sqrt{T}\mathbf{F}\mathbf{h}_t^k+\widehat{\mathbf{v}}_t^k\Vert_2^2 \}
	\end{aligned} \ .
	\end{equation}	
	
Solving Eq.~\ref{eq:g} directly is very difficult because of its complexity. So we decide to sample $\widehat{\mathbf{x}}_t$ across all $K$ channels in each pixel to simplify our formulation written by:
   \begin{equation}\small\label{eq:gt}
	\begin{aligned}
	\varGamma_j^*(\widehat{\mathbf{G}}_t)=&\mathop{arg \ min}\limits_{\varGamma_j(\hat{\mathbf{G}}_t)}\{\Vert\widehat{\mathbf{y}}_j-\varGamma_j(\widehat{\mathbf{X}}_t)^\top\varGamma_j(\widehat{\mathbf{G}}_t)\Vert_2^2\\
	+&\gamma\Vert \varGamma_j(\widehat{\mathbf{G}}_t)+\varGamma_j(\widehat{\mathbf{V}}_t)-\varGamma_j(\sqrt{T}\mathbf{F}\mathbf{H}_t)\Vert_2^2\\
	+&\theta_t\Vert\varGamma_j(\widehat{\mathbf{G}}_t)-\varGamma_j(\widehat{\mathbf{G}}_{t-1})\Vert_2^2 		
	\}
	\end{aligned}  \ ,
	\end{equation}
where $\varGamma_j(\widehat{\mathbf{X}})\in \mathbb{C}^{K\times 1} $ represents the vector containing values of all $K$ channels of $\widehat{\mathbf{X}}$ on pixel $j  (j=1,2,...,T)$. After derivation using Sherman Morrison formula, we can obtain its solution:
	\begin{equation}\small
	\varGamma_j^*(\widehat{\mathbf{G}}_t)=\frac{1}{\gamma+\theta_t}(\mathbf{I}-\frac{\varGamma_j(\widehat{\mathbf{X}}_t)\varGamma_j(\widehat{\mathbf{X}}_t)^\top}{\theta_t+\gamma+\varGamma_j(\widehat{\mathbf{X}}_t)^\top\nu_j(\widehat{\mathbf{X}}_t)})\boldsymbol{\rho} \ ,
	\end{equation}
where the vector $\boldsymbol{\rho}$ takes the form $\boldsymbol{\rho}=
\varGamma_j(\widehat{\mathbf{X}}_t)\widehat{\mathbf{y}}_j+\theta_t\varGamma_j(\hat{\mathbf{G}}_{t-1})-\gamma\varGamma_j(\widehat{\mathbf{V}}_t)+
\gamma\varGamma_j(\sqrt{T}\mathbf{F}{\mathbf{H}}_t)$ for presentation.

\textbf{Subproblem $\mathbf{H}$:} given $\theta_t,\widehat{\mathbf{G}}_t,\widehat{\mathbf{V}}_t$, we can optimize $\mathbf{h}^{k}$ by:
	\begin{equation}
	\small
	\mathop{arg \ min}\limits_{\mathbf{h}_k}\{{\frac{1}{2}\Vert \widetilde{\mathbf{u}}\odot \mathbf{h}_t^k\Vert_2^2+\frac{\gamma}{2}\Vert \widehat{\mathbf{g}}_t^k-\sqrt{T}\mathbf{F}\mathbf{h}_t^k+\widehat{\mathbf{v}}_t^k\Vert_2^2}\} \ .
	\end{equation}
	
The closed-form solution of $\mathbf{h}^{k}$ can be written by:
\begin{equation}\small
\begin{aligned}
\mathbf{h}^{k*}
=[\widetilde{\mathbf{U}}^\top\widetilde{\mathbf{U}}+\gamma T\mathbf{I}]^{-1}\gamma T (\mathbf{v}_t^k+\mathbf{g}_t^k) =
\frac{\gamma T(\mathbf{v}_t^k+\mathbf{g}_t^k)}{(\widetilde{\mathbf{u}}\odot \widetilde{\mathbf{u}})+\gamma T} \ ,
\end{aligned}
\end{equation}
where $\widetilde{\mathbf{U}}=diag (\widetilde{\mathbf{u}})\in \mathbb{R}^{T\times T}$ represents diagonal matrix.

\textbf{Subproblem \ $\theta_t$:} given other variables in Eq.~\ref{eq:8}, the optimal solution of $\theta_t$ can be determined as:
	\begin{equation}
	\small
	\begin{aligned}
	\theta_t^*=&\mathop{arg\ min}\limits_{\theta_t}\{\frac{\theta}{2}\sum_{k{}=1}^{K}\Vert\widehat{\mathbf{g}}_t^k-\widehat{\mathbf{g}}_{t-1}^k\Vert_2^2+\frac{1}{2}\Vert\theta_t-\tilde{\theta}\Vert_2^2\}\\
	=&\tilde{\theta}-\frac{\sum_{k=1}^{K}\Vert\widehat{\mathbf{g}}_t^k-\widehat{\mathbf{g}}_{t-1}^k\Vert_2^2}{2}
	\end{aligned} \ .
	\end{equation}

 \textbf{Lagrangian multiplier update:}
after solving three subproblems above, we can update Lagrangian multipliers as:
	\begin{equation}
	\small
	\widehat{\mathbf{V}}^{i+1}=\widehat{\mathbf{V}}^{i}+\gamma^{i}(\widehat{\mathbf{G}}^{i+1}-\widehat{\mathbf{H}}^{i+1}) \ ,
	\end{equation} 
where $i$ and $i+1$ denotes the iteration index and the step size regularization constant $\gamma$ (initially equals to $1$) takes the form of $\gamma^{(i+1)}$=$min(\gamma_{max},\beta \gamma^{i})$. ($\beta=10$,  $\gamma_{max}=10000$)

By iteratively solving the four subproblems above, we can optimize our objective function effectively and obtain the optimal filter $\widehat{\mathbf{G}}_t$ and temporal regularization parameter $\theta_t$ in frame $t$. Then  $\widehat{\mathbf{G}}_t$ is used for detection in frame $t+1$.

\subsection{Object Localization}
The tracked object is localized by searching for the maximum value of response map $\mathcal{R}_t$ calculated by:
\begin{small}
\begin{equation}\small
\mathcal{R}_t=\mathscr{F}^{-1}\sum_{k=1}^{K}(\widehat{\mathbf{z}}^k_t\odot \widehat{\mathbf{g}}^k_{t-1}),
\end{equation} 	
\end{small}

\noindent where $\mathcal{R}_t$ is the response map in frame $t$, $\mathscr{F}^{-1}$ denotes the inverse Fourier transform (IFT) operator and $\widehat{\mathbf{z}}^k_t$ represents the Fourier form of extracted feature map in frame $t$.

\section{Localization by Tracking}
Self-localization for UAV is essential for autonomous navigation. To develop a robust and universal localization system in dynamic and uncertain environments, we introduce visual object tracking into UAV localization for the first time. Specifically, we utilize the open-source software in~\cite{Faessler2014ICRA},  but employ AutoTrack to track four objects simultaneously instead of segmenting LEDs in the infrared spectrum. The main work-flow is briefly described below.

\textbf{Prerequisites:} the system requires the knowledge of four object configuration (non-symmetric), \ie, their positions in the world coordinate (observed in motion capture system), and intrinsic UAV-mounted camera parameters.


\textbf{Initialization and tracking:} after manually assigning four objects, AutoTrack starts to track them independently and output their location in the RGB image. Different to the system~\cite{Faessler2014ICRA} only applicable in infrared spectrum, our system can be used in versatile environments.

\textbf{Correspondence search and pose optimization:} correspondence between the tracked object configuration in the world coordinate and tracked results in image frames is firstly clarified, then the final 6D pose is optimized by fine-tuning the reprojection error~\cite{Faessler2014ICRA}.
\section{Experiments}
In this section, we firstly evaluate the tracking performance of AutoTrack with current state-of-the-art trackers on four difficult UAV benchmarks~\cite{Li2017AAAI,Du2018ECCV,Mueller2016ECCV,Wen2018visdrone}. Then, the proposed localization system is evaluated on
Quanser\footnote{\url{https://www.quanser.com/products/autonomous-vehicles-research-studio/}} platform in the indoor practical scenarios. The experiments of tracking performance evaluation are conducted using MATLAB R2018a on a PC with an i7-8700K processor (3.7GHz), 32GB RAM and NVIDIA GTX 2080 GPU. The tests of localization system are run on ROS~\cite{Quigley2009ros} using C++. For the hyper parameters of AutoTrack, we set $\delta=0.2$, $\nu=2\times10^{-5}$,  $\zeta=13$. The threshold of $\phi$ is 3000, ADMM iteration is set to 4. The sensitivity analysis of all the parameters can be found in the supplementary material.

\subsection{Evaluation on UAV Datasets}
For rigorous  and comprehensive evaluation,  the comparison between AutoTrack with the state-of-the-art methods is reported on four challenging and authoritative UAV benchmarks: DTB70~\cite{Li2017AAAI}, UAVDT~\cite{Du2018ECCV}, UAV123@10fps~\cite{Mueller2016ECCV} and VisDrone2018-test-dev~\cite{Wen2018visdrone}, with a total number of 119,830 frames. Noted that we use the same evaluation criteria with the four benchmarks~\cite{ Li2017AAAI, Du2018ECCV,Mueller2016ECCV, Wen2018visdrone}.

\subsubsection{Comparison with deep-based trackers}
\begin{figure}[t]
	\begin{center}
		\subfigure[] { 
			\begin{minipage}{0.226\textwidth}
				\centering
				\includegraphics[width=1\columnwidth]{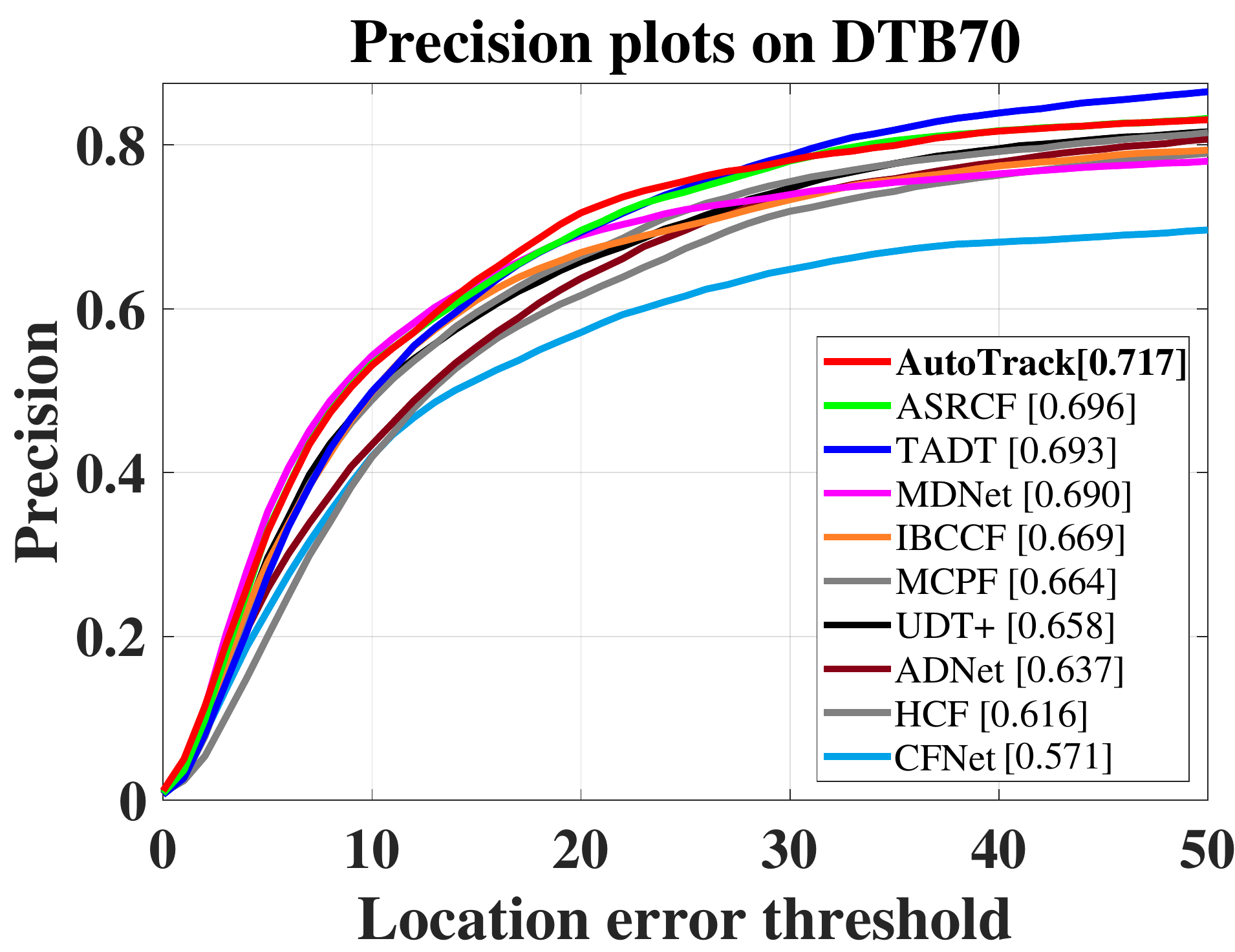}
			\end{minipage}
		}
		\subfigure[] { 
			\begin{minipage}{0.226\textwidth}
				\centering
				\includegraphics[width=1\columnwidth]{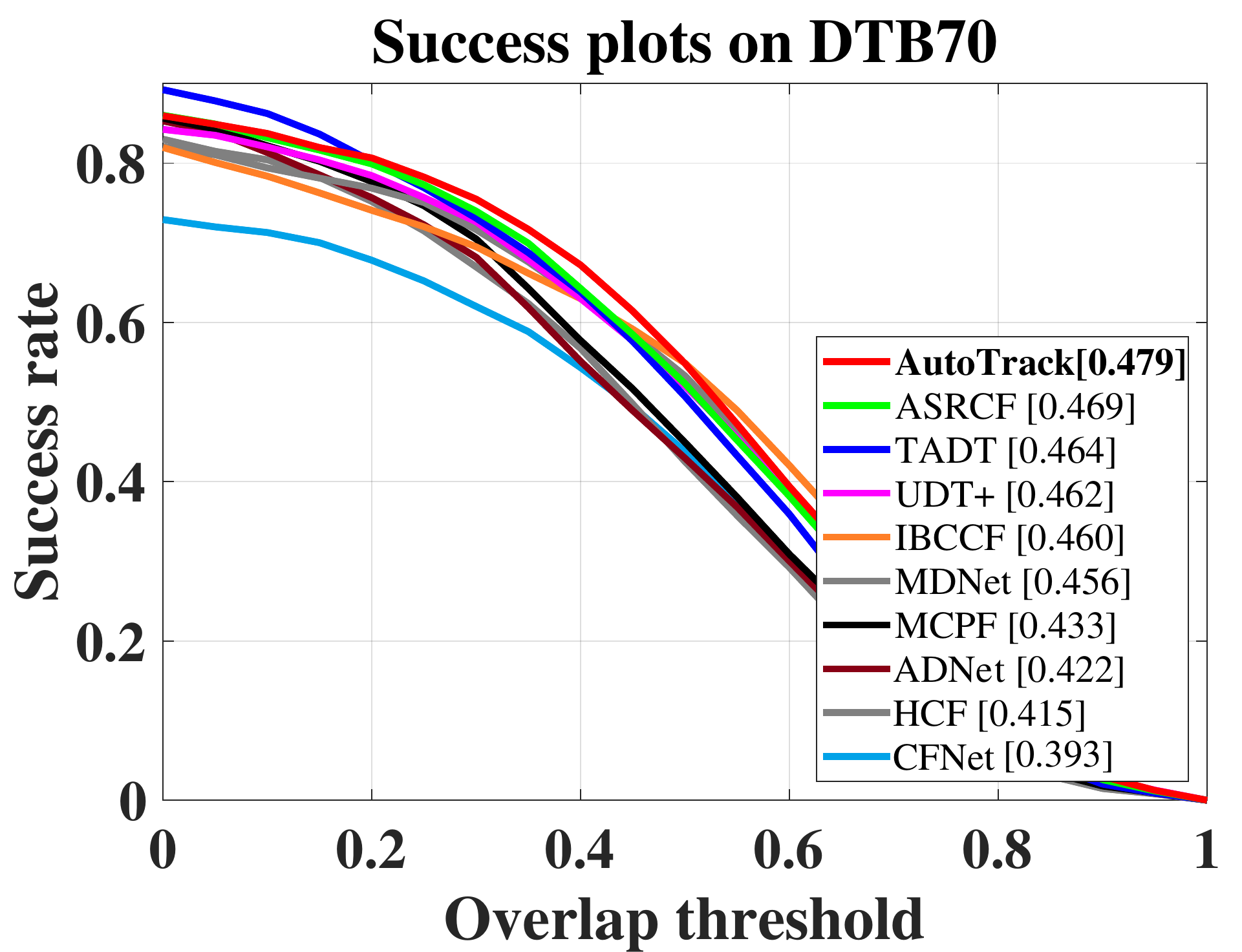}
			\end{minipage}
		}
	\end{center}
	\caption{Overall comparison with deep trackers on DTB70~\cite{Li2017AAAI}. AutoTrack ranks first place in both precision and success rate.}
	\label{fig:DTB70_deep}
\end{figure}

\ \ \ \textbf{DTB70:} DTB70~\cite{Li2017AAAI}, composed of 70 difficult UAV image sequences, primarily addresses the problem of severe UAV motion. In addition, various cluttered scenes and objects with different sizes as well as aspect ratios are included. We compare AutoTrack with nine state-of-the-art deep trackers, \ie, 
ASRCF~\cite{Dai2019CVPR}, TADT~\cite{Li2019TADT}, HCF~\cite{Ma2015ICCV}, ADNet~\cite{Yun2017CVPR}, CFNet~\cite{Valmedre2017CVPR}, UDT+~\cite{WangNing2019CVPR},  IBCCF~\cite{li2017integrating}, MDNet~\cite{Nam2016CVPR}, MCPF~\cite{Zhang2017CVPR}, on DTB70, and the final results are reported in Fig.~\ref{fig:DTB70_deep}. Only with hand-crafted features, AutoTrack outperforms deep feature-based trackers (ASRCF~\cite{Dai2019CVPR}, HCF~\cite{Ma2015ICCV}, MCPF~\cite{Zhang2017CVPR} and IBCCF~\cite{li2017integrating}) and  pre-trained deep architecture-based trackers, \ie, MDNet~\cite{Nam2016CVPR}, ADNet~\cite{Yun2017CVPR}, UDT+~\cite{WangNing2019CVPR} and CFNet~\cite{Valmedre2017CVPR}. In summary, AutoTrack exhibits strong robustness against drastic UAV motion without losing efficiency, and also demonstrates a generality in tracking different objects in various scenes.

\textbf{UAVDT:} UAVDT~\cite{Du2018ECCV} mainly emphasizes vehicle tracking in various scenarios. Weather condition, flying altitude and camera view are three categories addressed by UAVDT. Compared to deep trackers including ASRCF~\cite{Dai2019CVPR}, TADT~\cite{Li2019TADT}, SiameseFC~\cite{Bertinetto2016ECCV}, DSiam~\cite{Guo2017learning}, MCCT~\cite{Wang2018CVPR}, ADNet~\cite{Yun2017CVPR}, CFNet~\cite{Valmedre2017CVPR}, DeepSTRCF~\cite{Li2018CVPR}, UDT+~\cite{WangNing2019CVPR}, HCF~\cite{Ma2015ICCV}, C-COT~\cite{Danelljan2016ECCV}, ECO~\cite{Danelljan2017CVPR}, IBCCF~\cite{li2017integrating}, MCPF~\cite{Zhang2017CVPR} and CREST~\cite{Song2017ICCV}, AutoTrack with a single CPU exhibits the best performance in terms of precision and speed, as shown in Table~\ref{tab:UAVDT_deep}. In a word, AutoTrack has extraordinary performance in vehicle tracking despite omnipresent challenges.
\begin{table}[b]
	\setlength{\tabcolsep}{1.3mm}
	\footnotesize
	\centering
	\caption{Precision and speed comparison between AutoTrack with deep trackers on UAVDT~\cite{Du2018ECCV}. * means GPU speed.  \textcolor[rgb]{ 1,  0,  0}{Red}, \textcolor[rgb]{ 0,  1,  0}{green} and \textcolor[rgb]{ 0,  0,  1}{blue} respectively mean the first, second and third place.}
	\vspace{0.2cm}
	\begin{tabular}{ccc||ccc}
		\hline
		Tracker&Precision&FPS&Tracker&Precision&FPS\\
		\hline\hline
		\textbf{AutoTrack}&\textcolor[rgb]{ 1,  0,  0}{\textbf{71.8}}&\textcolor[rgb]{ 1,  0,  0}{\textbf{65.4}}&UDT+\cite{WangNing2019CVPR} & 69.7&\textcolor[rgb]{ 0,  1,  0}{\textbf{60.4*}} \\
		DeepSTRCF\cite{Li2018CVPR}&66.7&6.6*&ADNet\cite{Yun2017CVPR}&68.3&7.6*\\
		DSiam\cite{Guo2017learning}&\textcolor[rgb]{ 0,  1,  0}{\textbf{70.4}}&15.9* &TADT\cite{Li2019TADT}&67.7&32.5*\\
		MCPF\cite{Zhang2017CVPR}&66.0& 0.67* &MCCT\cite{Wang2018CVPR}&67.1&8.6* \\
		Siamese\cite{Bertinetto2016ECCV}&68.1& 37.9* &ECO\cite{Danelljan2017CVPR}&\textcolor[rgb]{ 0,  0,  1}{\textbf{70.0}}&16.4* \\
		C-COT\cite{Danelljan2016ECCV}&65.6& 1.1* &CREST\cite{Song2017ICCV}&64.9&4.3* \\
		ASRCF\cite{Dai2019CVPR}&\textcolor[rgb]{ 0,  0,  1}{\textbf{70.0}}& 24.1* &HCF\cite{Ma2015ICCV}&60.2&20.15* \\
		CFNet\cite{Valmedre2017CVPR}&68.0 & \textcolor[rgb]{ 0,  0,  1}{\textbf{41.1*}} &IBCCF\cite{li2017integrating}&60.3&3.39* \\
		\hline
	\end{tabular}%
	\label{tab:UAVDT_deep}%
\end{table}%
\subsubsection{Comparison with CPU-based trackers}
\begin{figure*}[t]
	\begin{center}

		\subfigure[] { \label{fig:DTB70} 
			\begin{minipage}{0.236\textwidth}
				\centering
				\includegraphics[width=1\columnwidth]{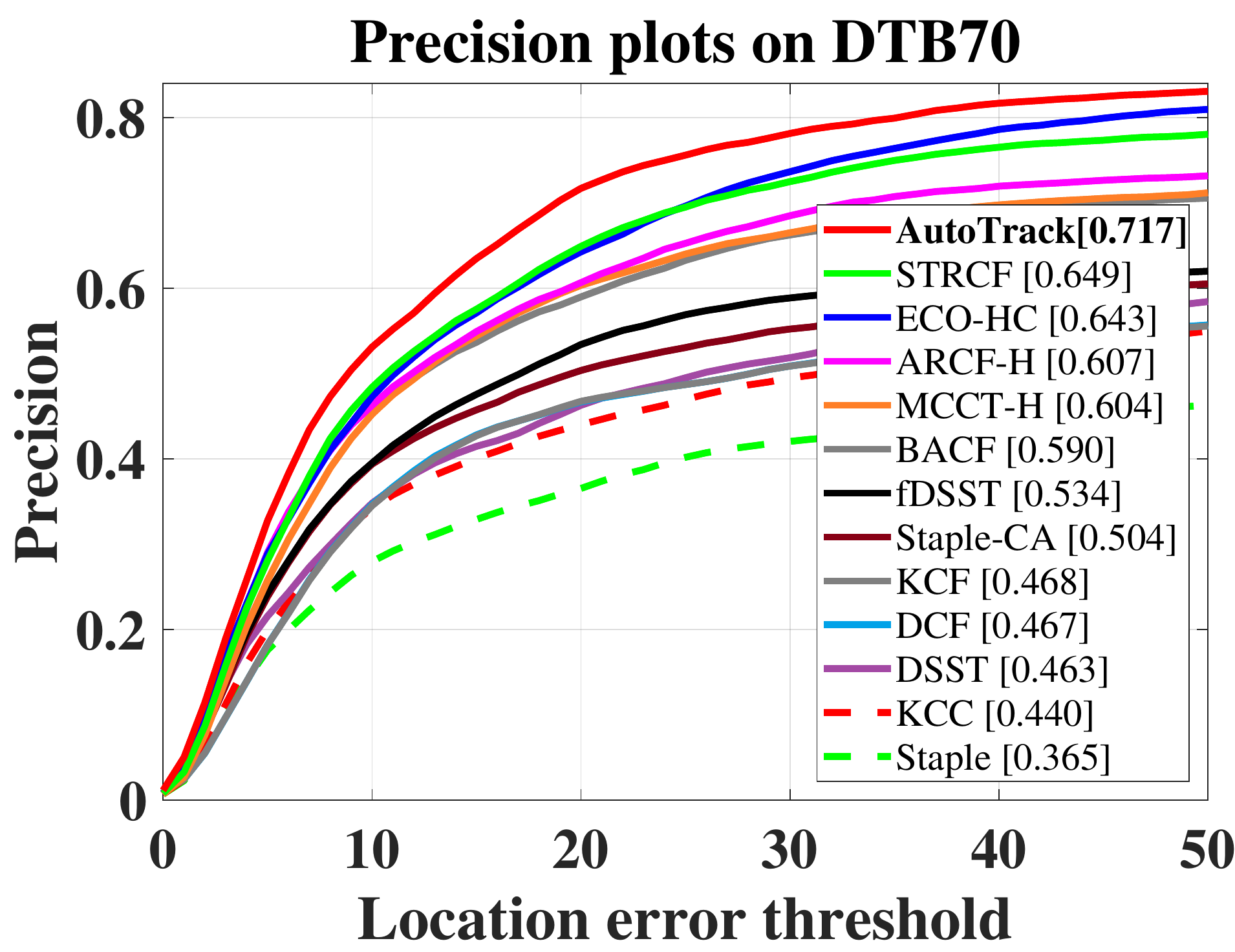}
				\\
				\includegraphics[width=1\columnwidth]{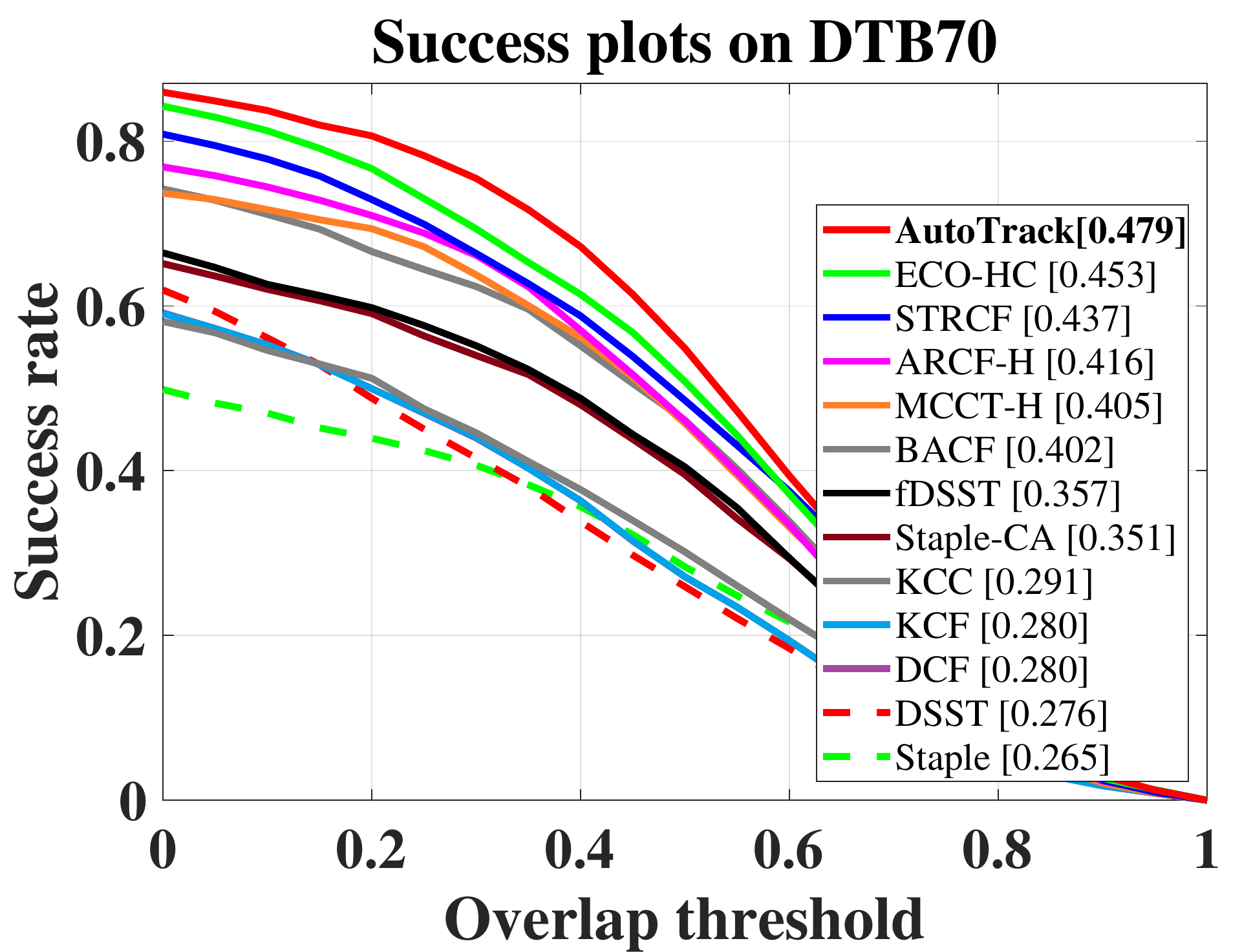}
			\end{minipage}
		}
		\subfigure[] { \label{fig:UAVDT} 
			\begin{minipage}{0.236\textwidth}
				\centering
				\includegraphics[width=1\columnwidth]{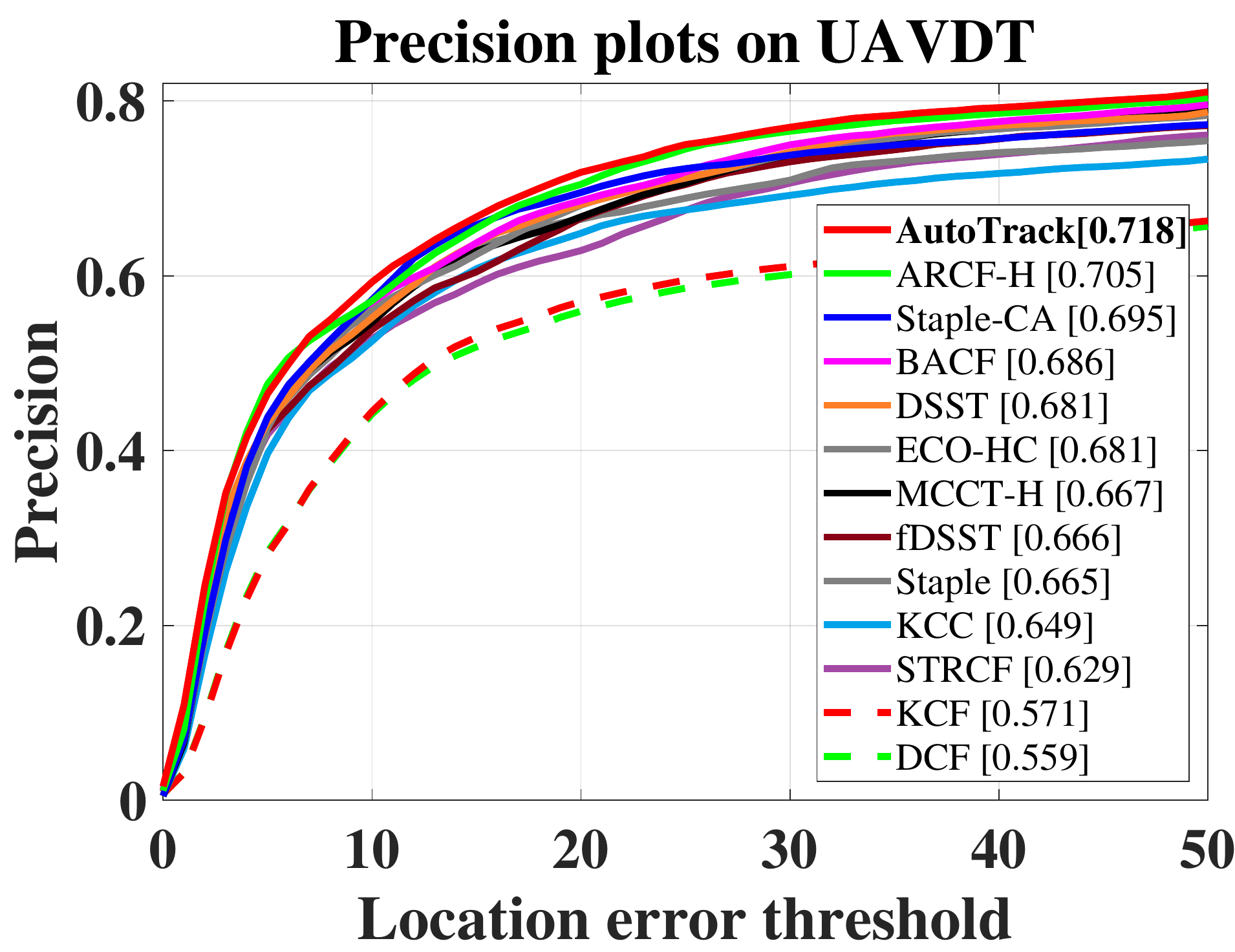}
				\\
				\includegraphics[width=1\columnwidth]{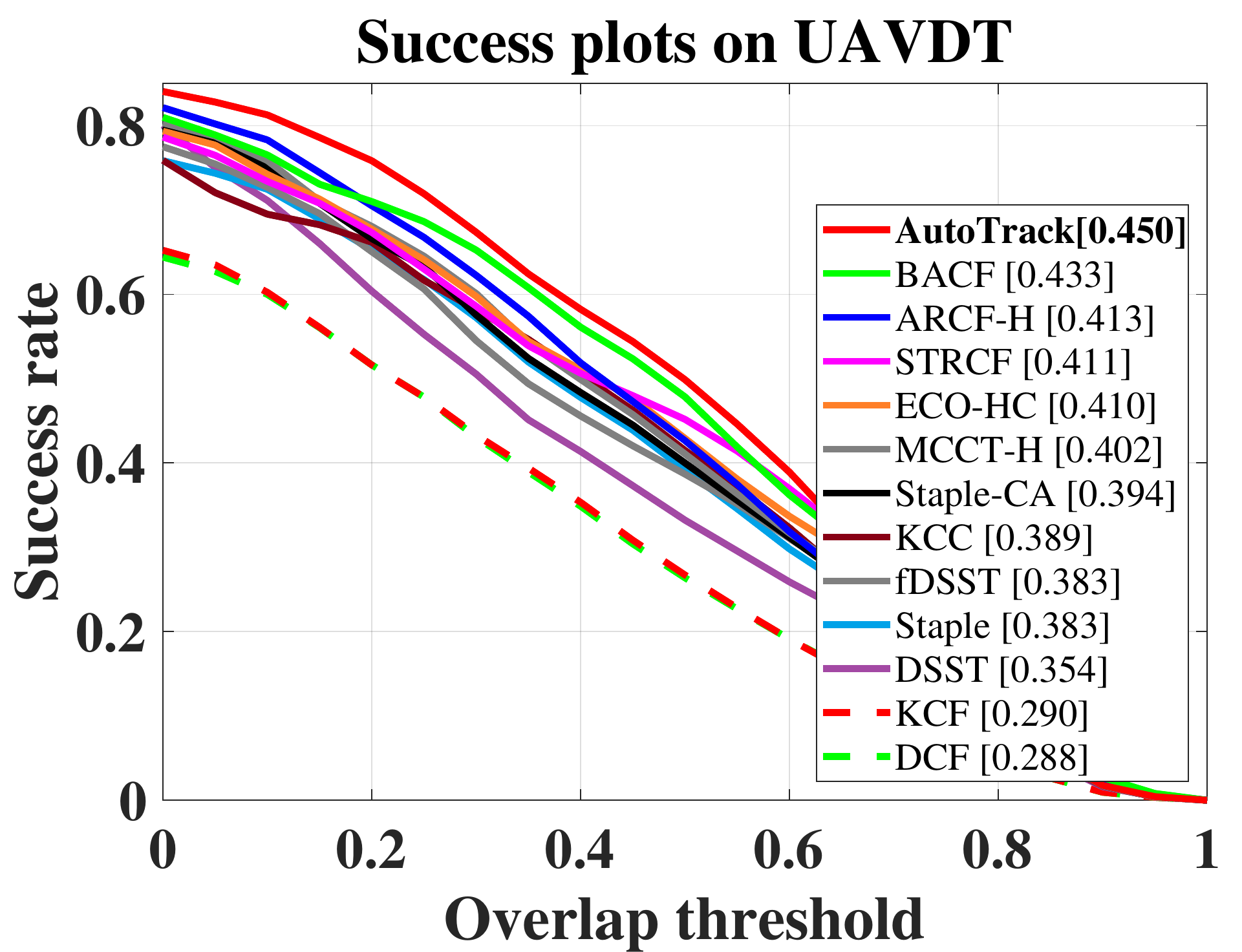}
			\end{minipage}
		}
		\subfigure[] { \label{fig:UAV123} 
			\begin{minipage}{0.236\textwidth}
				\centering
				\includegraphics[width=1\columnwidth]{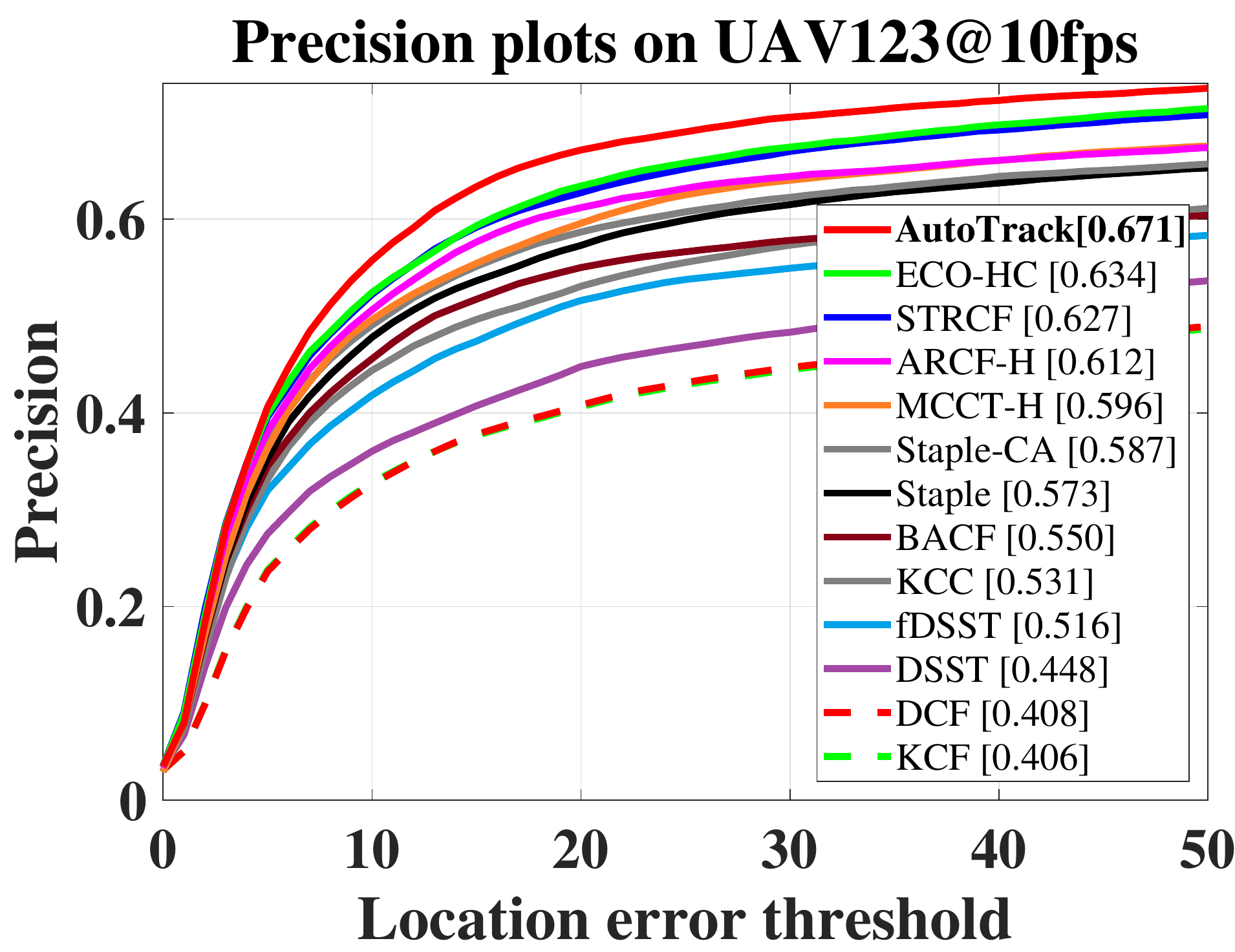}
				\\
				\includegraphics[width=1\columnwidth]{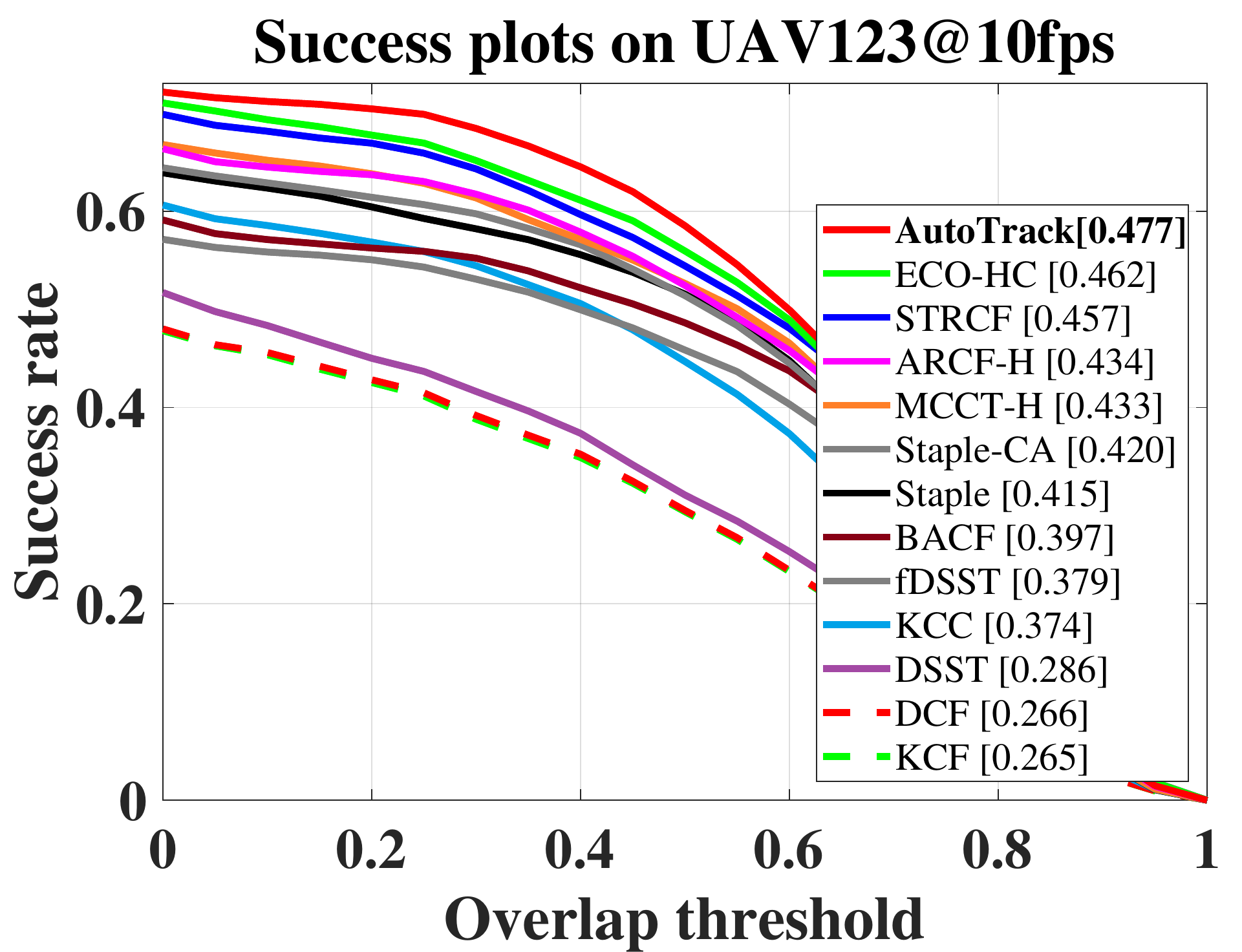}
			\end{minipage}
		}
		\subfigure[] { \label{fig:VisDrone} 
			\begin{minipage}{0.236\textwidth}
				\centering
				\includegraphics[width=1\columnwidth]{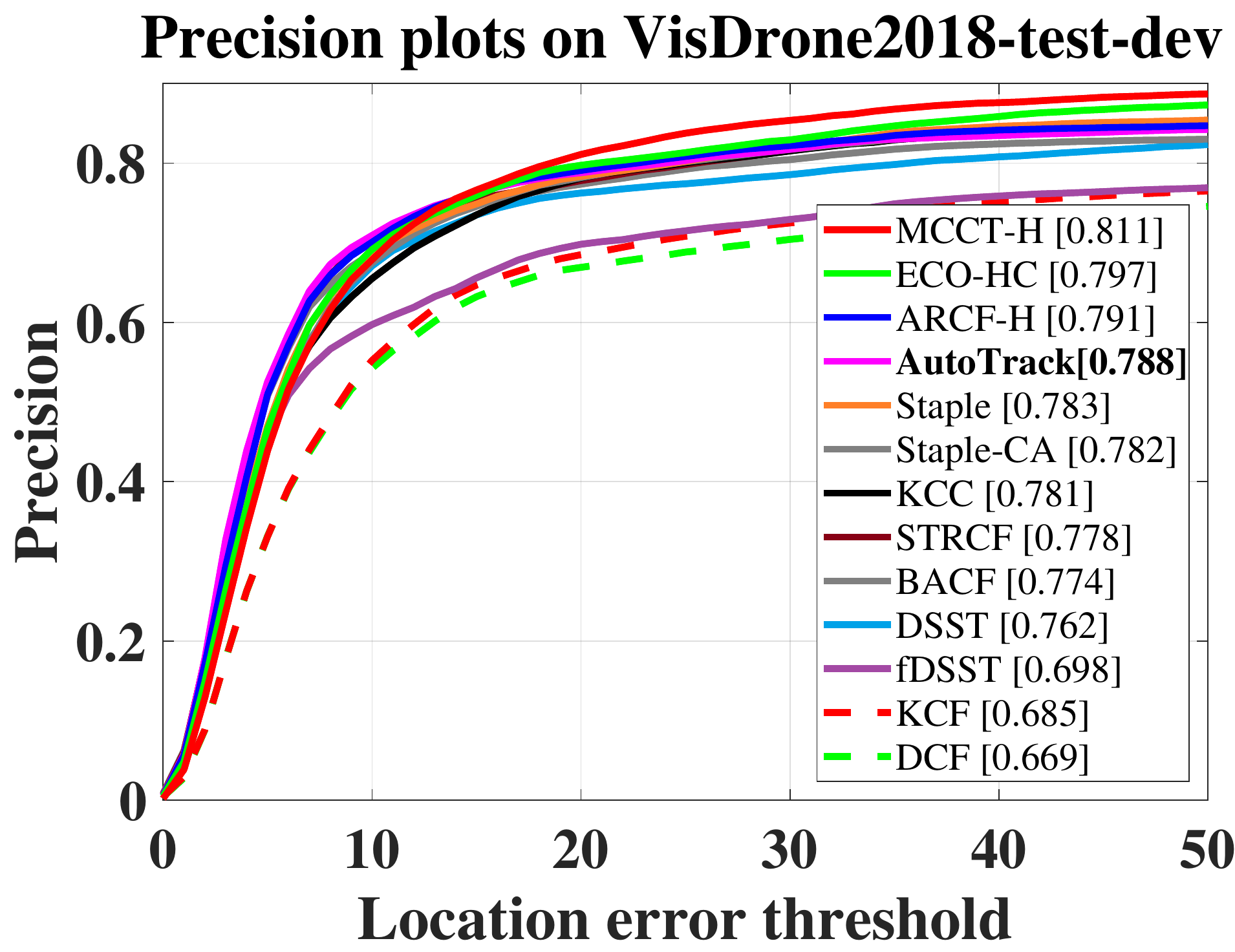}
				\\
				\includegraphics[width=1\columnwidth]{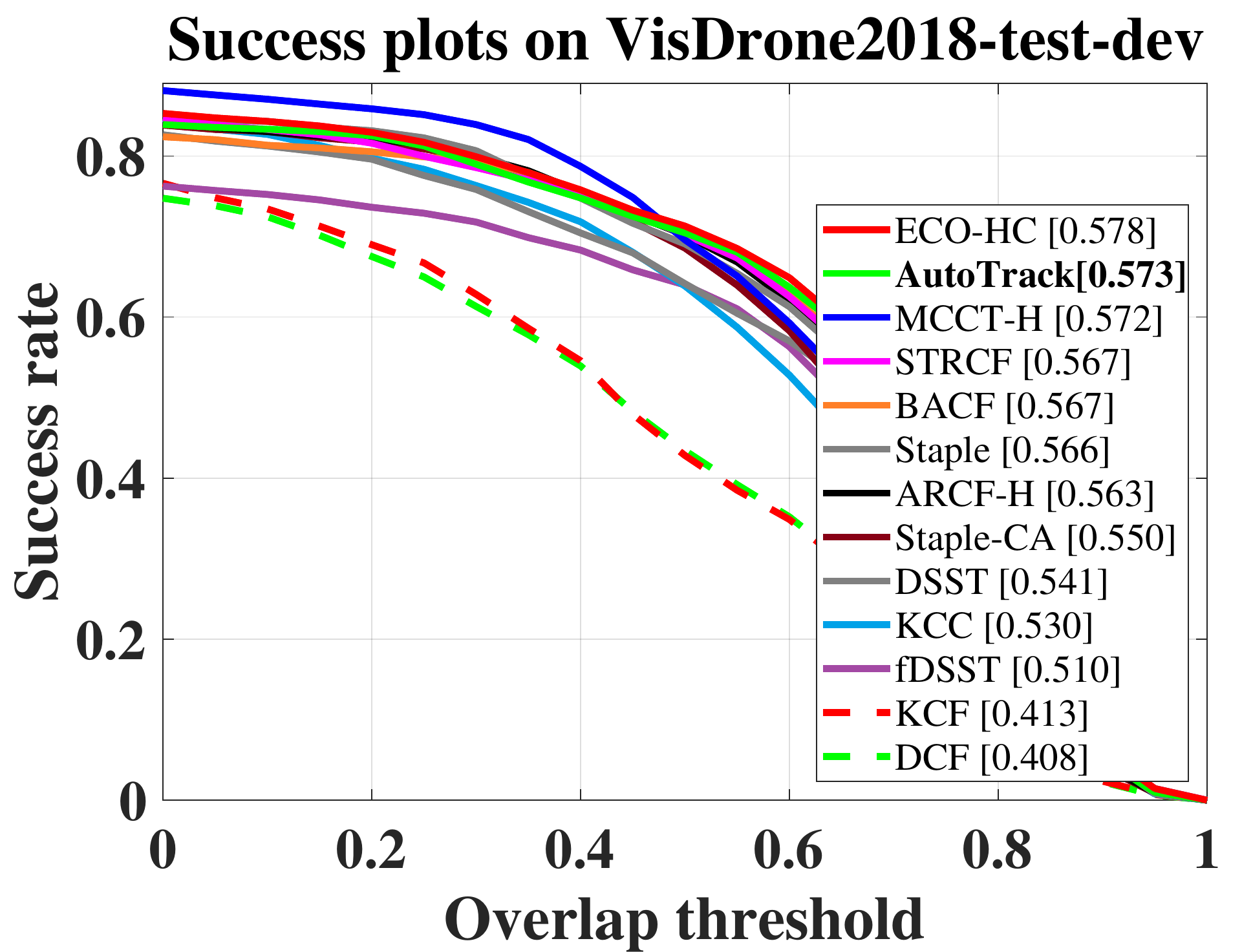}
			\end{minipage}
		}
	\end{center}
	
	\caption{Overall performance of CPU-based real-time trackers on (a) DTB70~\cite{Li2017AAAI} (b) UAVDT~\cite{Du2018ECCV} (c) UAV123@10fps~\cite{Mueller2016ECCV} and (d) VisDrone2018-test-dev~\cite{Wen2018visdrone}. Two measures for one-pass evaluation (OPE)~\cite{wu2013online} are used for evaluation. Precision plot can demonstrate the percentage of scenarios when the distance between estimated bounding box and ground truth one is smaller than different thresholds, and the score at 20 pixels is used for ranking. Success plot can display the percentage of situations when the overlap between estimated bounding box and ground truth one is greater than different thresholds. Area under curve (AUC) is utilized for ranking.}
	\label{fig:overall}
\end{figure*}

\begin{figure}[b]
	\begin{center}
		\includegraphics[width=0.99\columnwidth]{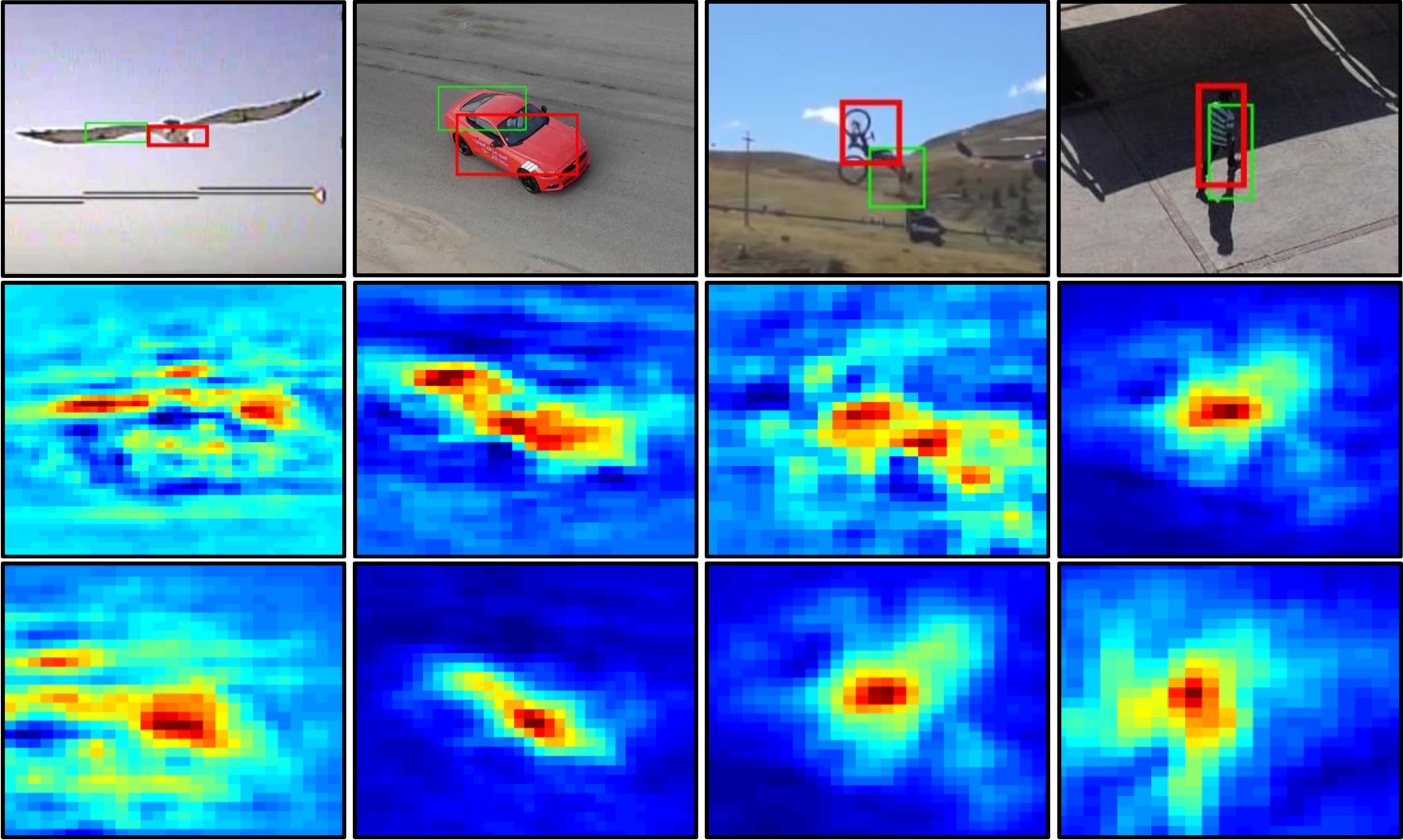}
		\caption{Tracking results and response maps of AutoTrack (red box) and STRCF (green box) of \textit{bird1\_3}, \textit{car18}, \textit{MountainBike5} and \textit{person12\_2}. AutoTrack (third row) has less distraction in response than STRCF (second row) due to automatic regularization.}
		\label{fig:ablation}
	\end{center}
\end{figure}

\begin{figure}[b]
	\begin{center}
		\includegraphics[width=1.0\columnwidth]{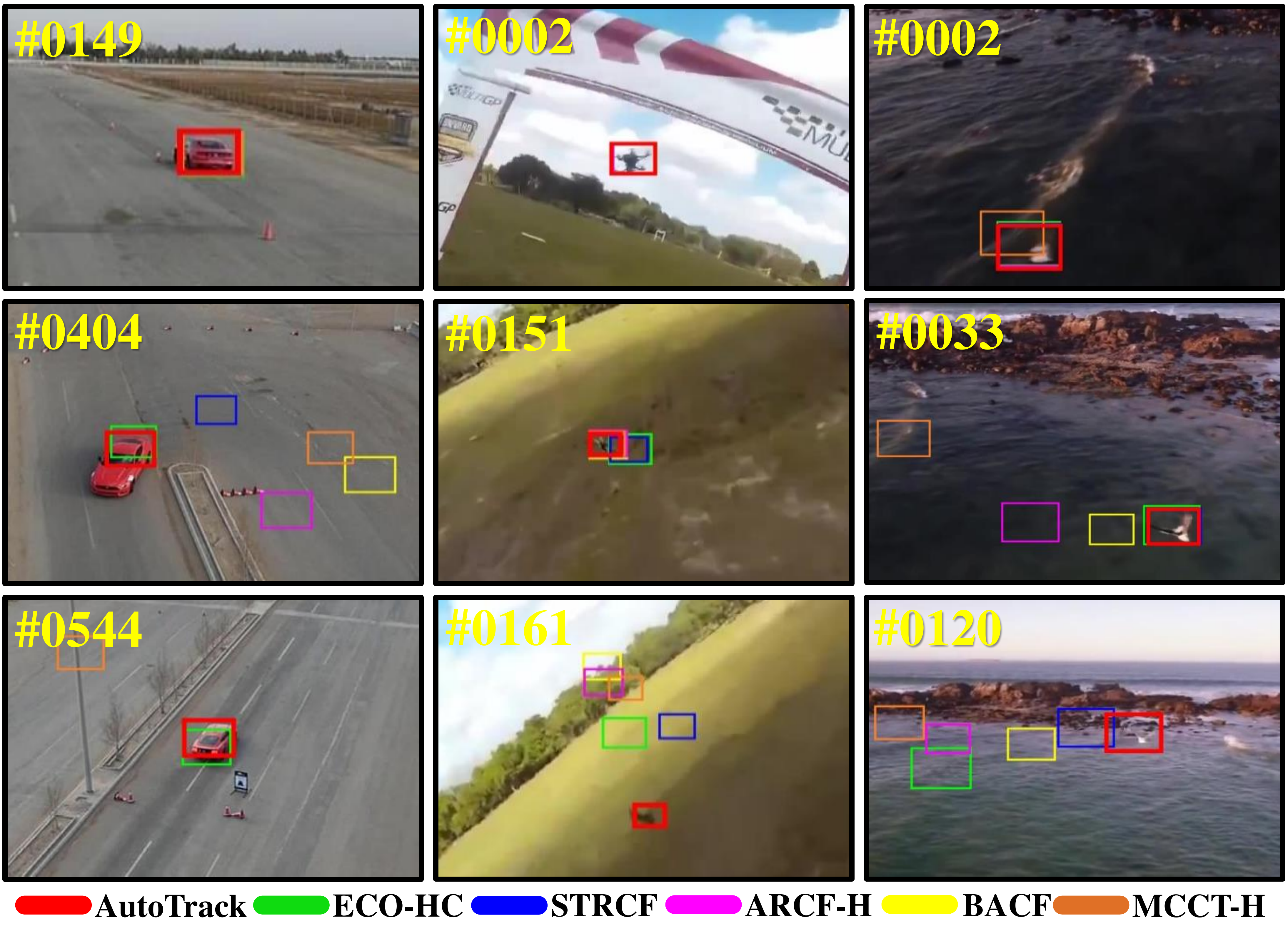}
		\caption{Screenshots of \textit{Car16\_2}, \textit{ChasingDrones}, and \textit{Gull1}.}
		\label{fig:tracking_results}
	\end{center}
\end{figure}

\ \  Twelve real-time trackers (with a speed of $>$$30$fps), \ie, KCF~\cite{Henriques2015PAMI}, DCF~\cite{Henriques2015PAMI}, KCC~\cite{wang2018kernel} fDSST~\cite{Danelljan2017PAMI}, DSST~\cite{Danelljan2014BMVA}, BACF~\cite{Kiani2017ICCV}, STAPLE-CA~\cite{Mueller2017CVPR}, STAPLE~\cite{Mueller2017CVPR}, MCCT-H~\cite{Wang2018CVPR}, STRCF~\cite{Li2018CVPR}, ECO-HC~\cite{Danelljan2017CVPR}, ARCF-H~\cite{Huang2019ICCV}, and five non-real-time ones, \ie,  SRDCF~\cite{Danelljan2015ICCV},
SAMF~\cite{Li2014ECCV}, CSR-DCF~\cite{LukezicCVPR2017}, SRDCFdecon~\cite{Danelljan2016CVPR}, ARCF-HC~\cite{Huang2019ICCV} are used for comparison. The results of real-time trackers on four datasets are displayed in Fig.~\ref{fig:overall}. Besides, the average performance of top ten CPU-based trackers in terms of speed and precision is demonstrated in the Table~\ref{tab:average_performance}. It can be seen that AutoTrack is the best real-time tracker on CPU. Some tracking results are demonstrated in Fig.~\ref{fig:ablation} and Fig.~\ref{fig:tracking_results}.
\begin{center}
	\begin{table*}[!t]
		\footnotesize
		\setlength{\tabcolsep}{0.7mm}
		\centering
		\caption{Average speed (fps) and precision of top ten CPU-based trackers on four benchmarks. \textcolor[rgb]{ 1,  0,  0}{Red}, \textcolor[rgb]{ 0,  1,  0}{green} and \textcolor[rgb]{ 0,  0,  1}{blue} respectively mean the first, second and third place. All the reported speed is run on a single CPU. Noted that AutoTrack is the best real-time tracker on CPU.}
		\vspace{0.2cm}
		\begin{tabular}{c c c c c c c c c c c }
			\hline
			{Tracker}&
			\textbf{AutoTrack}&{ARCF-HC}\cite{Huang2019ICCV}&{ECO-HC}\cite{Danelljan2017CVPR}&{ARCF-H}\cite{Huang2019ICCV}&{STRCF}\cite{Li2018CVPR}&{MCCT-H}\cite{Wang2018CVPR}&{STAPLE\_CA}\cite{Mueller2017CVPR}&{BACF}\cite{Kiani2017ICCV}&{CSR-DCF}\cite{LukezicCVPR2017}&{SRDCF}\cite{Danelljan2015ICCV}\\ \hline \hline
			{Precision} & \textcolor[rgb]{1, 0, 0}{\textbf{72.4}}&\textcolor[rgb]{0,  1,  0}{\textbf{71.9}}  & \textcolor[rgb]{0,  0,  1}{\textbf{69.1}}& {67.3} &{67.1}&{67.0}&{64.2}&{65.6}&{67.7}&{62.7}\\ 
			\hline
			{Speed} & \textcolor[rgb]{0, 1, 0}{\textbf{59.2}}&{19.3} &  \textcolor[rgb]{1, 0, 0}{\textbf{69.5}}& {53.4}  &{28.4}&\textcolor[rgb]{0, 0, 1}{\textbf{58.8}}&{58.5}&{53.1}&{11.8}&{14.2}\\ 
			\hline
		\end{tabular}%
		\label{tab:average_performance}%
	\end{table*}%
\end{center}

\textbf{Overall performance evaluation:}  AutoTrack has outperformed all the CPU-based real-time trackers in both precision and success rate on DTB70~\cite{Li2017AAAI}, UAVDT~\cite{Du2018ECCV} and UAV123@10fps~\cite{Mueller2016ECCV}. On VisDrone2018-test-dev~\cite{Wen2018visdrone}, AutoTrack achieves comparable performance with the best tracker MCCT-H and ECO-HC in terms of precision and success rate. 
As for the average performance of top ten CPU-based trackers, AutoTrack has the best performance in precision with the second fast speed of 59.2fps, only slower than ECO-HC (69.5fps), however, we have achieved an average improvement of 4.8\% in precision compared to ECO-HC.
Moreover, AutoTrack has an advantage of 7.9\% in precision and 108.5\% in speed over the baseline STRCF.
\begin{figure*}[t]
	\begin{center}

		\subfigure[] { \label{fig:att1} 
			\begin{minipage}{0.236\textwidth}
				\centering
				\includegraphics[width=1\columnwidth]{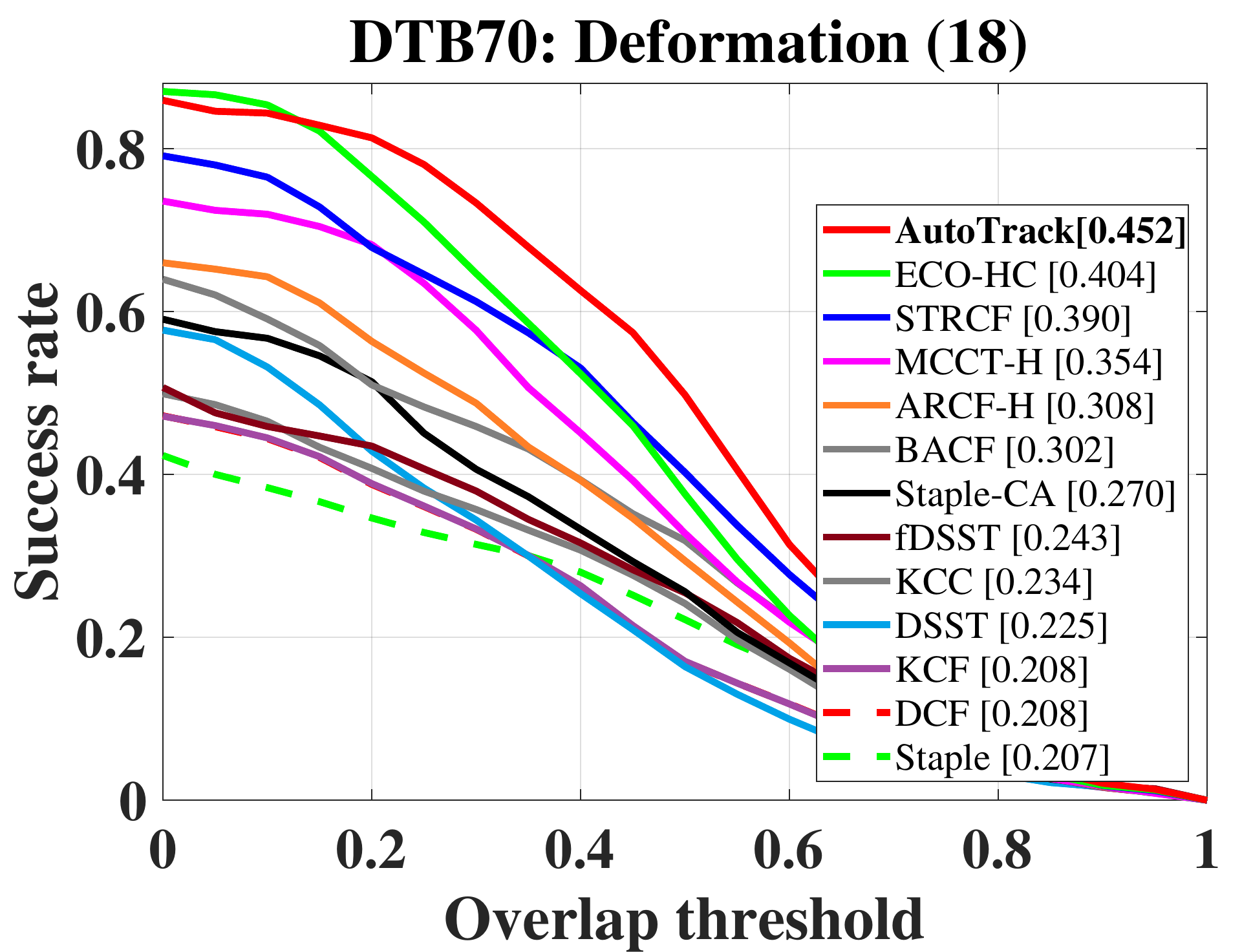}
			\end{minipage}
		}
		\subfigure[] { \label{fig:att2} 
			\begin{minipage}{0.236\textwidth}
				\centering
				\includegraphics[width=1\columnwidth]{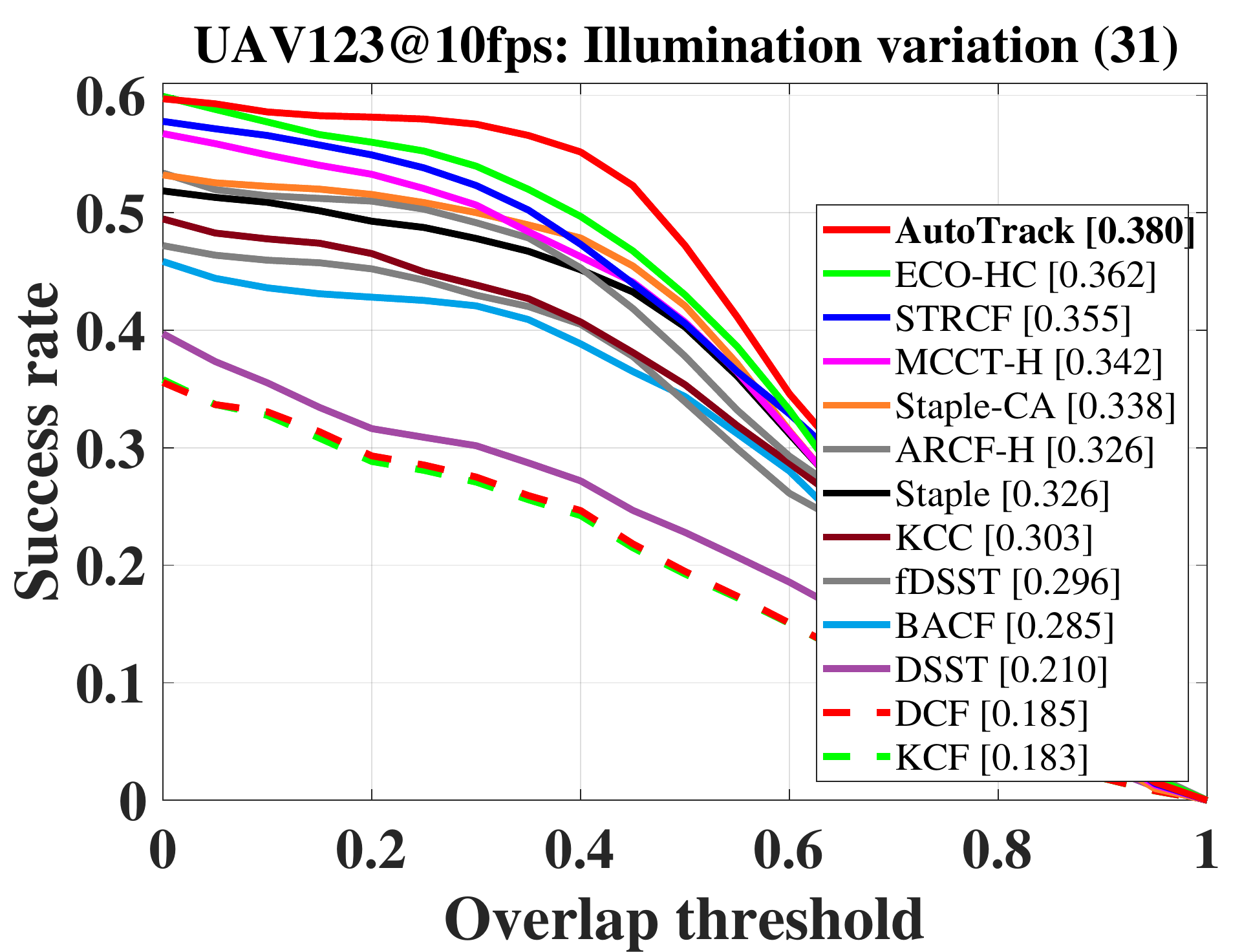}
			\end{minipage}
		}
		\subfigure[] { \label{fig:att3} 
			\begin{minipage}{0.236\textwidth}
				\centering
				\includegraphics[width=1\columnwidth]{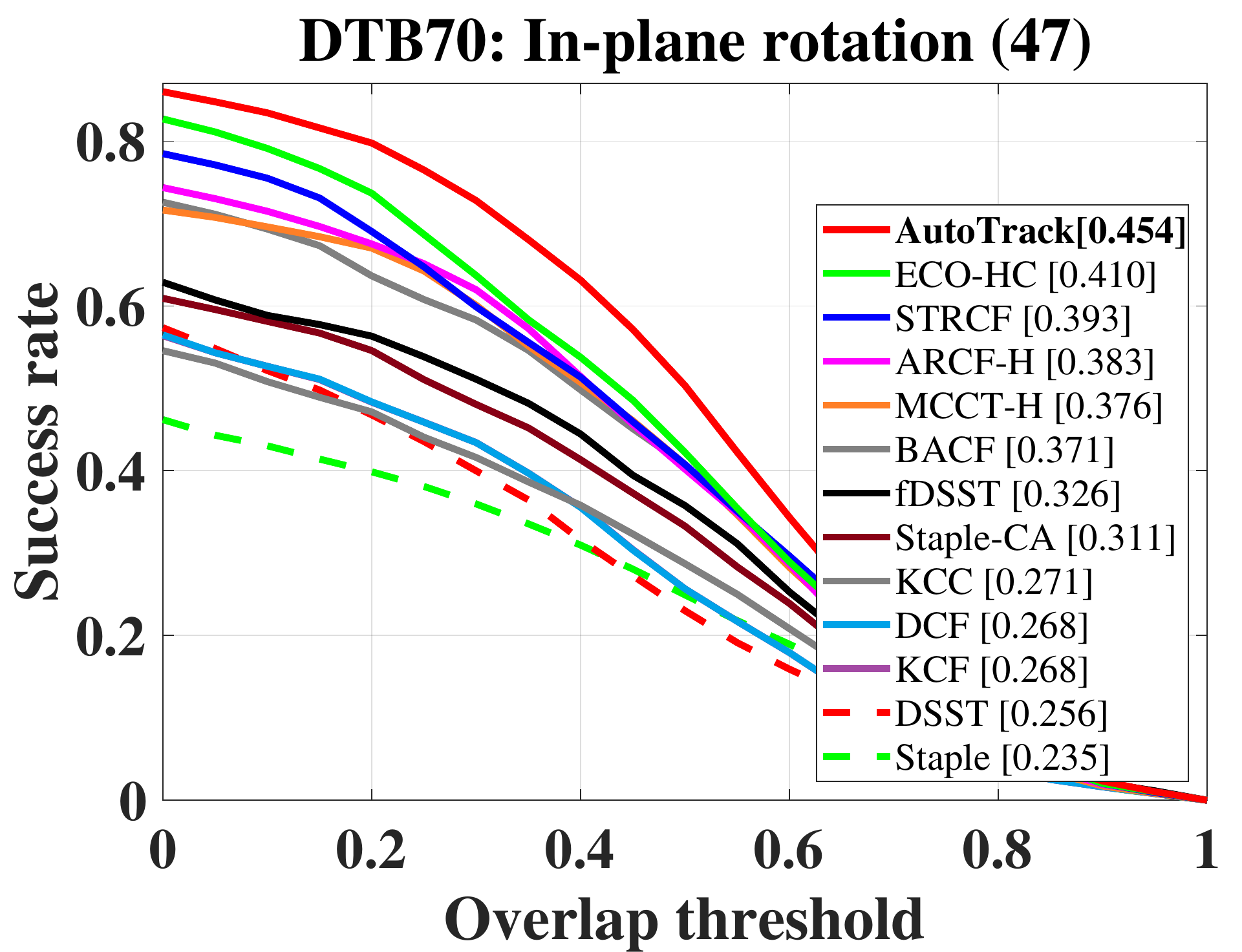}
			\end{minipage}
		}
		\subfigure[] { \label{fig:att4} 
			\begin{minipage}{0.236\textwidth}
				\centering
				\includegraphics[width=1\columnwidth]{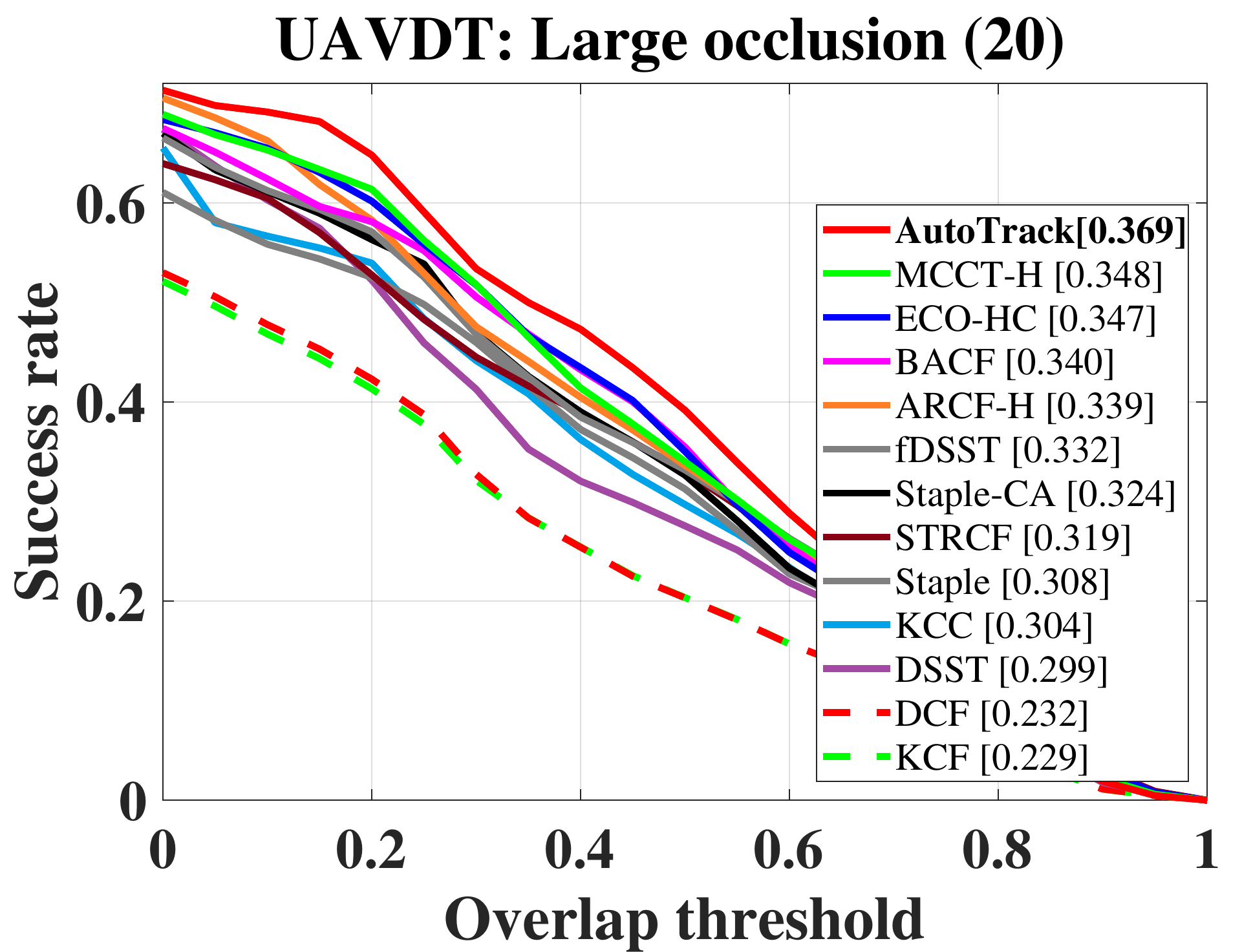}
			\end{minipage}
		}
		
		\subfigure[] { \label{fig:att5} 
			\begin{minipage}{0.236\textwidth}
				\centering
				\includegraphics[width=1\columnwidth]{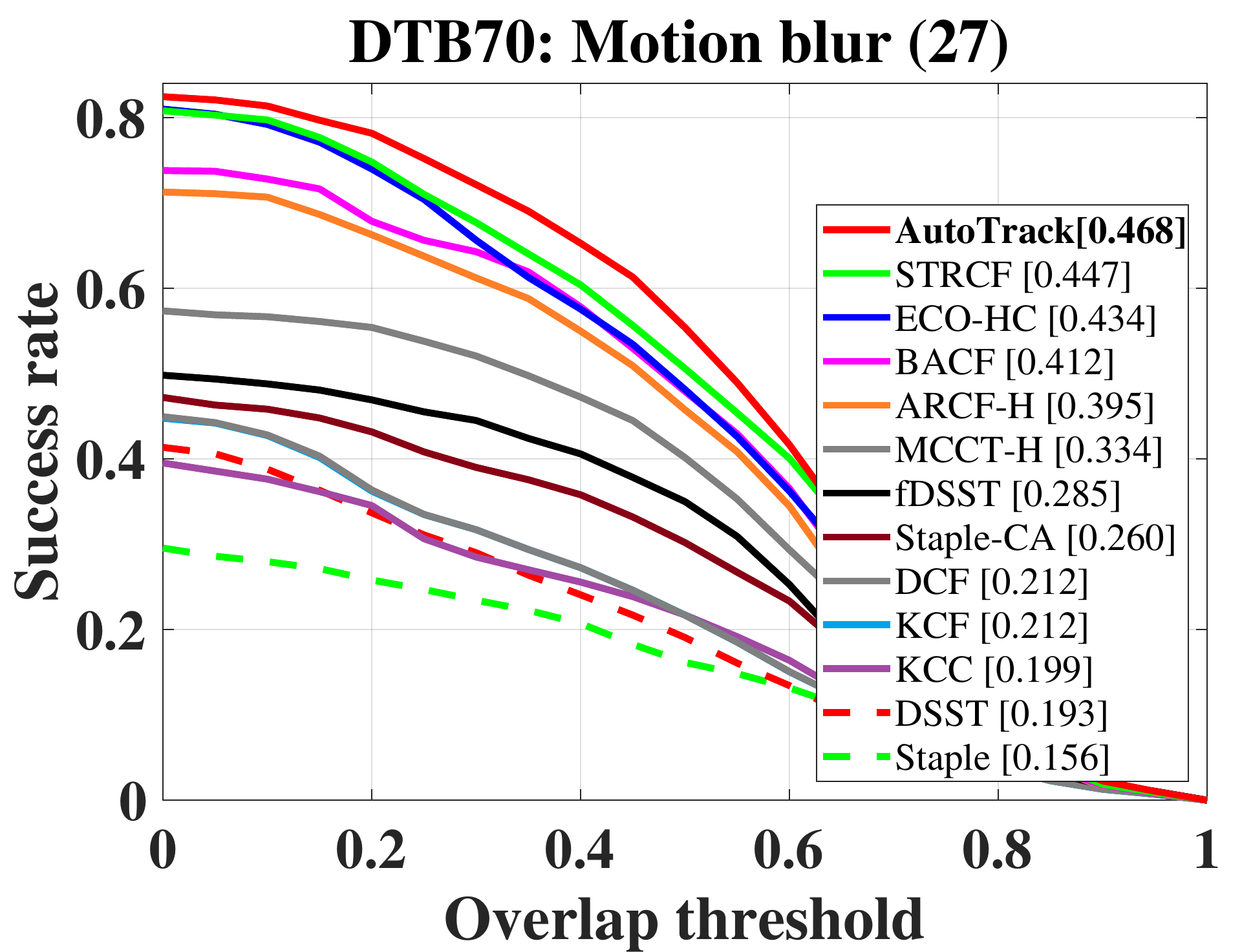}
			\end{minipage}
		}
		\subfigure[] { \label{fig:att6} 
			\begin{minipage}{0.236\textwidth}
				\centering
				\includegraphics[width=1\columnwidth]{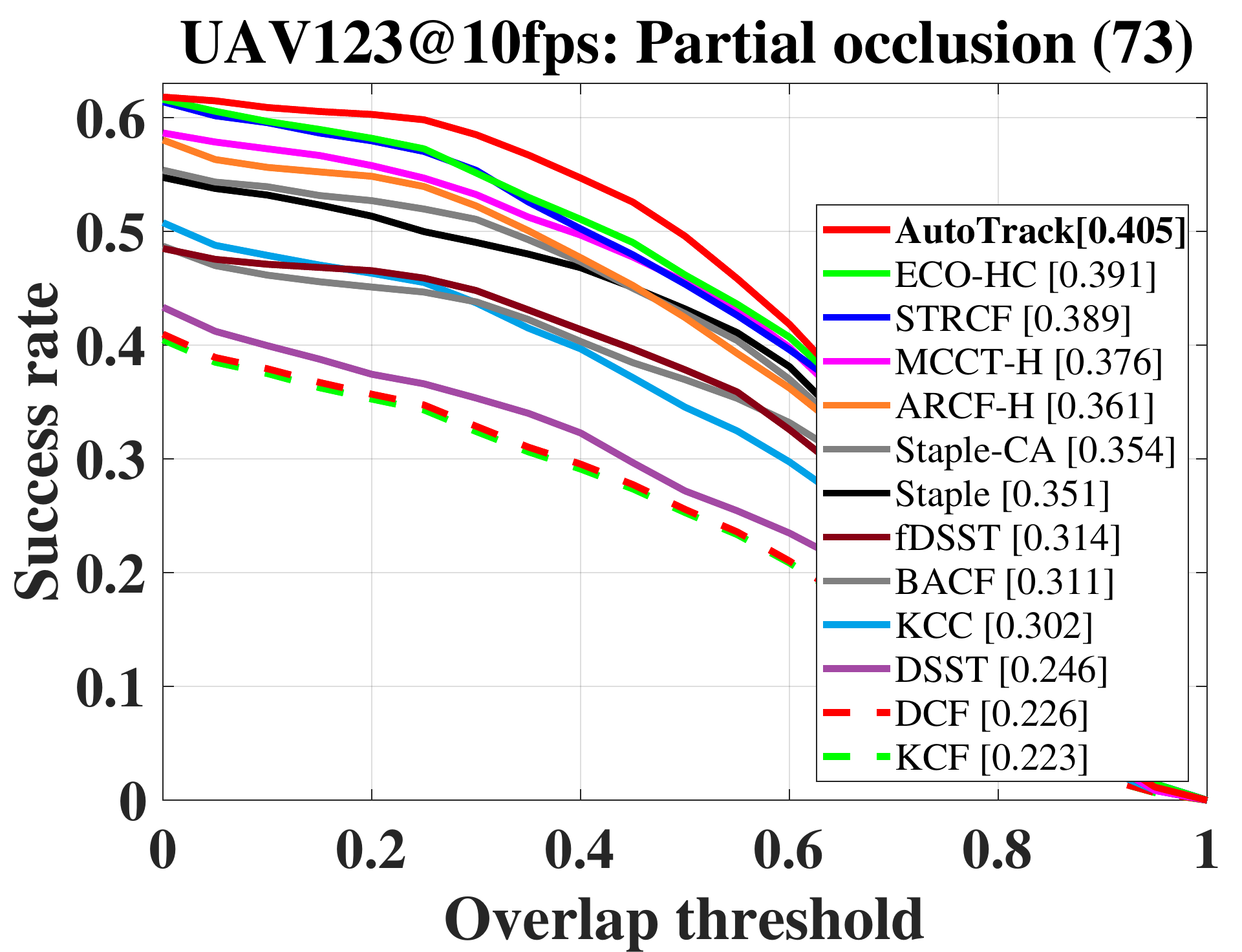}
			\end{minipage}
		}
		\subfigure[] { \label{fig:at7} 
			\begin{minipage}{0.236\textwidth}
				\centering
				\includegraphics[width=1\columnwidth]{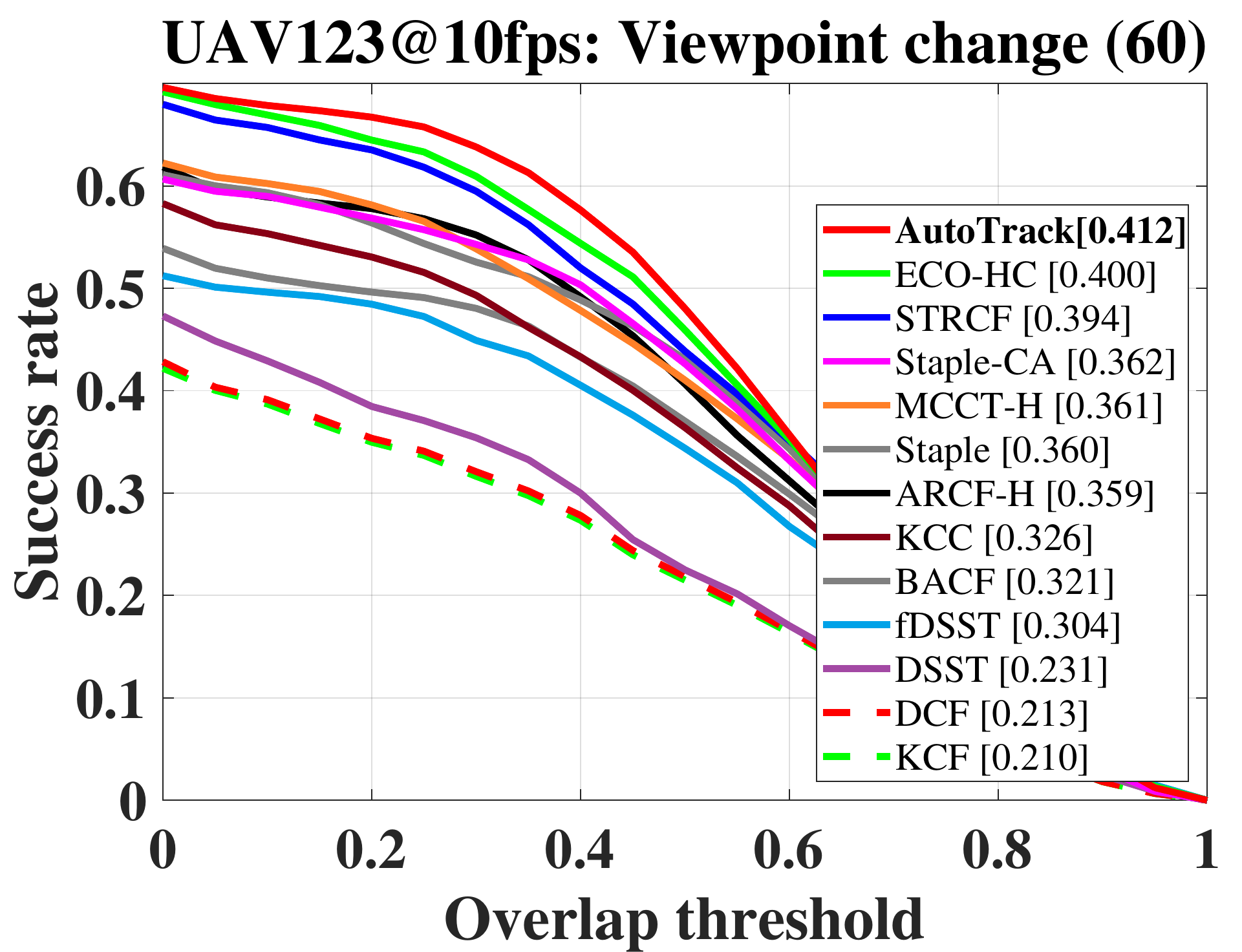}
			\end{minipage}
		}
		\subfigure[] { \label{fig:att8} 
			\begin{minipage}{0.236\textwidth}
				\centering
				\includegraphics[width=1\columnwidth]{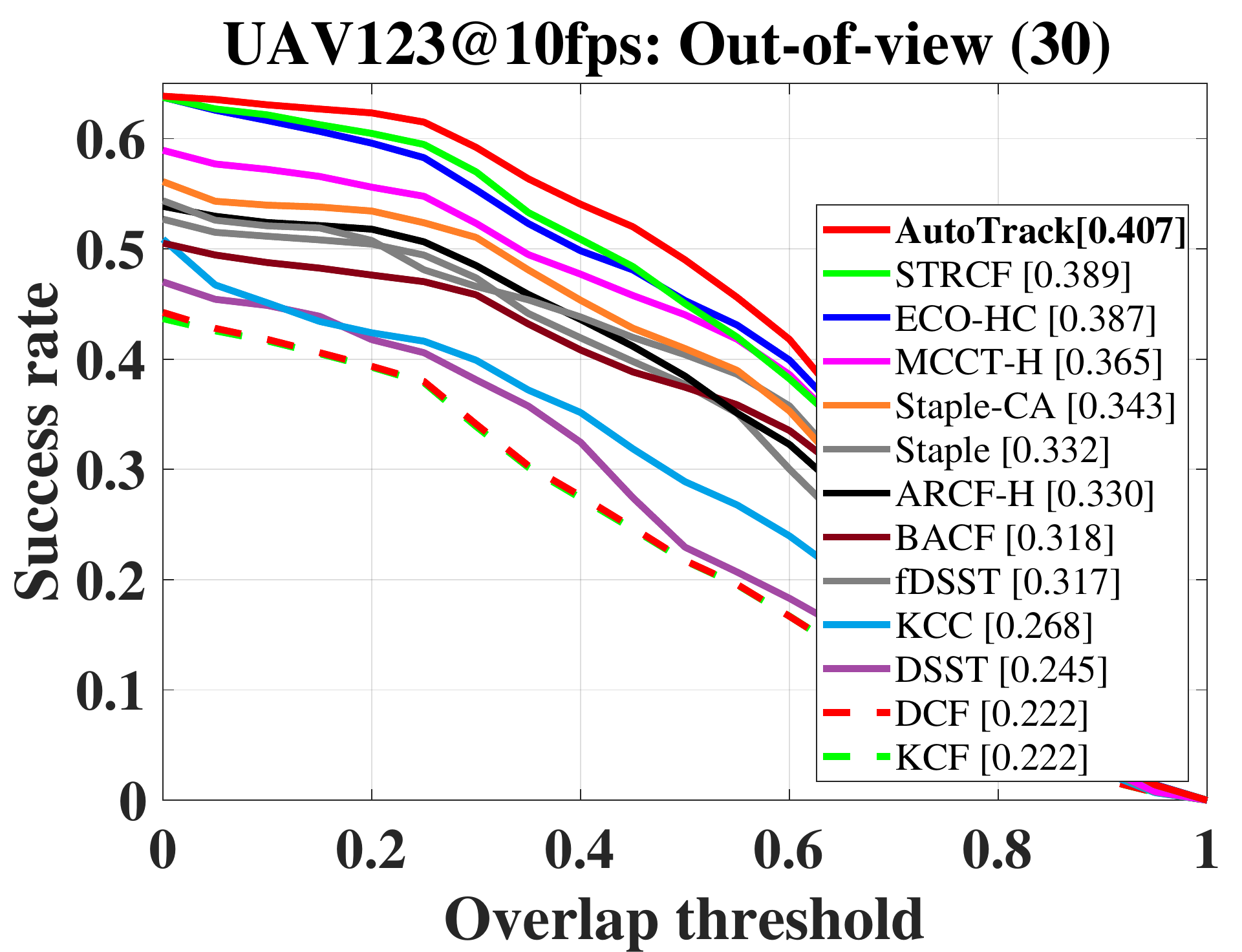}
			\end{minipage}
		}
	\end{center}
	
	\caption{Attribute-based comparison on deformation, illumination variation, in-plane rotation, large occlusion, motion blur, partial occlusion, viewpoint change, and out-of-view. More attribute-based evaluations can be seen in the supplementary material.}
	\label{fig:attribute}
\end{figure*}

\noindent\textbf{Remark 2:} M. Muller~\etal created a 10fps dataset from the recorded 30fps one~\cite{Mueller2016ECCV}, thus the movement of tracked object between successive frames is larger, bringing more challenges. On UAV123@10fps, AutoTrack achieves a remarkable advantage of 5.8\% in precision than the second best ECO-HC, proving its robustness against large motion. 

\noindent\textbf{Remark 3:} Compared to ARCF-HC~\cite{Huang2019ICCV} solely repressing the global response variation using a fixed parameter, we fully utilize the local-global information to fine-tune the spatio-temporal regularization term in an automatic manner. Extensive experiments have shown that AutoTrack achieves better performance while providing a much faster speed which is 3.1 times that of ARCF-HC. 

\textbf{Attribute-based evaluation:}  Success plots of eight  attributes are exhibited in Fig.~\ref{fig:attribute}. In the normal appearance change scenarios (deformation, in-plane-rotation, viewpoint change), AutoTrack improves STRCF by 15.9\%, 15.5\% and 4.6\% in success rate because the automatic temporal regularization can smoothly help filter adapt to new appearance. In illumination variation and large occlusion (aberrant appearance variation), AutoTrack has a superiority of 7.0\% and 15.7\% compared to STRCF in light of adaptive spatial regularization as well as aberrance monitoring mechanism which can stop training before contamination.
\subsubsection{Ablation study}
\label{sec:aberrance_repression_evaluation}
\ \ \ To validate the effectiveness of our method, AutoTrack is compared to itself with different modules enabled. The overall evaluation is presented in Table~\ref{tab:effectivenssstudy}. With each module (automatic spatial regularization ASR, automatic temporal regularization ATR) added to the STRCF, the performance is smoothly improved. It is noted that ATR can also bring a gain in speed compared to ASR because we can reduce meaningless and detrimental training on contaminated samples. In addition, response maps of some frames are illustrated in Fig.~\ref{fig:ablation}. It can be clearly seen that response of our method is more reliable than that of baseline.
\begin{table}[t]
	\setlength{\tabcolsep}{4.5mm}
	\small
	\centering
	\caption{Ablation study of AutoTrack. ASR and ATR respectively represents automatic spatial and temporal regularization.}
	\vspace{0.2cm}
	\begin{tabular}{cc cc c}
		\hline
		Tracker&Precision&AUC&FPS\\
		\hline\hline
		STRCF&0.671&0.468&28.4  \\
		STRCF + ASR&0.716&0.489&53.7\\
		STRCF + ATR&0.714&0.492&60.0 \\
		AutoTrack&0.724&0.495&59.2\\
		\hline
	\end{tabular}%
	\label{tab:effectivenssstudy}%
\end{table}%
\begin{figure}[b]
	\begin{center}
		\includegraphics[width=\columnwidth]{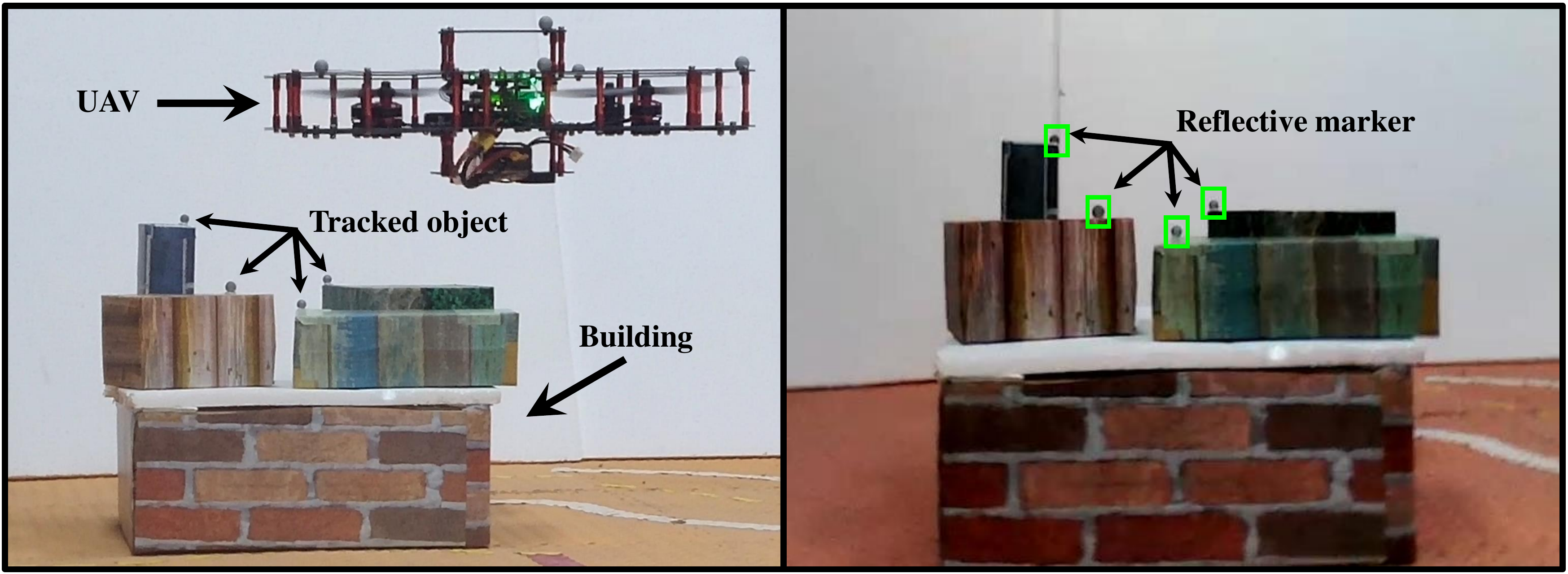}
		\caption{Experiment setup (left) and view from the UAV-mounted camera (right). The tracked objects (reflective markers whose ground truth locations are known in Quanser motion-capture system) for UAV localization are denoted as four green rectangles.}
		\label{fig:localization}
	\end{center}
\end{figure}
\subsection{Evaluation of Localization System}
\begin{figure*}[t]
	\begin{center}

		\subfigure[] { 
			\begin{minipage}{0.32\textwidth}
				\centering
				\includegraphics[width=1\columnwidth]{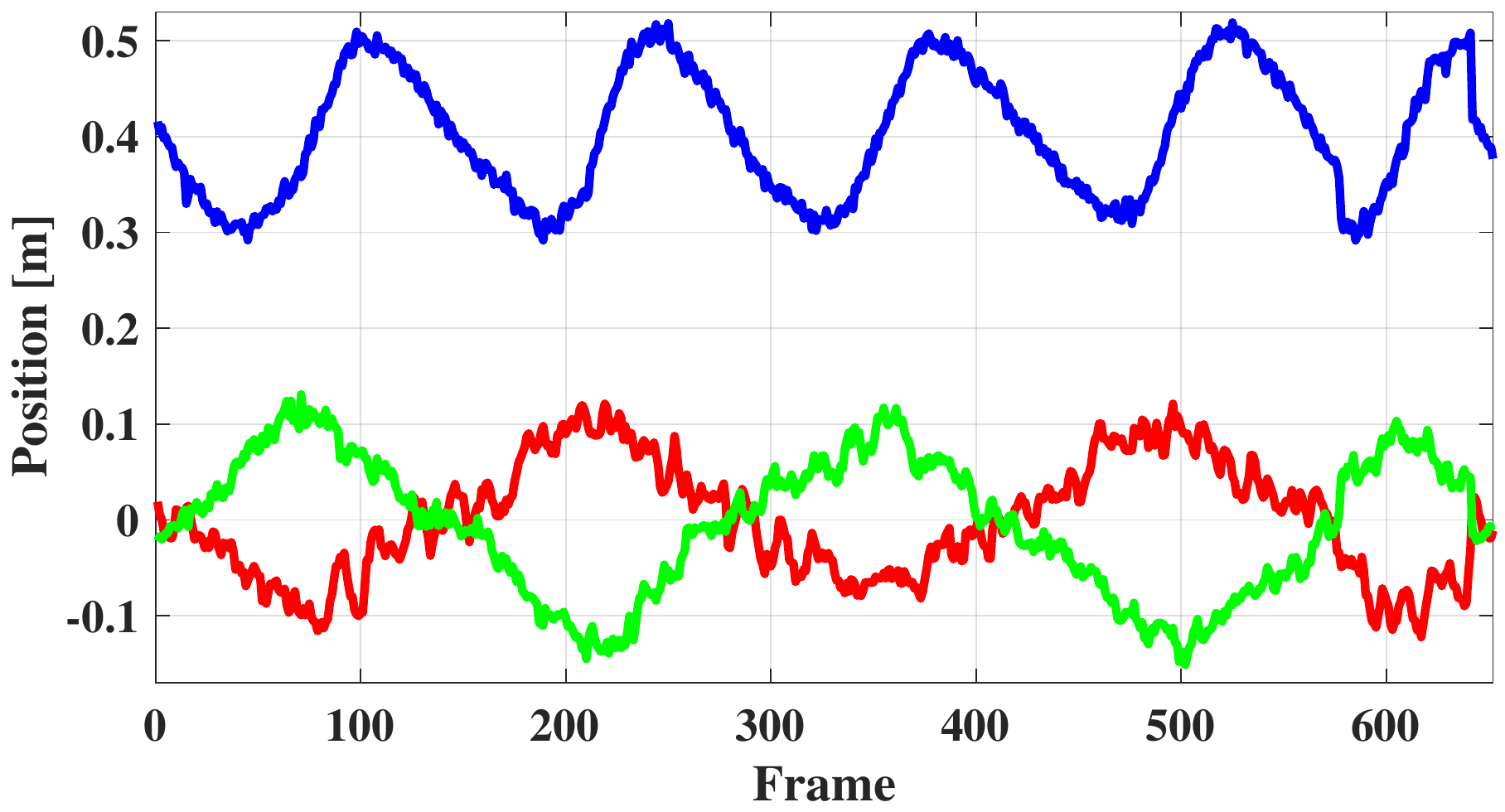}
				\\
				\includegraphics[width=1\columnwidth]{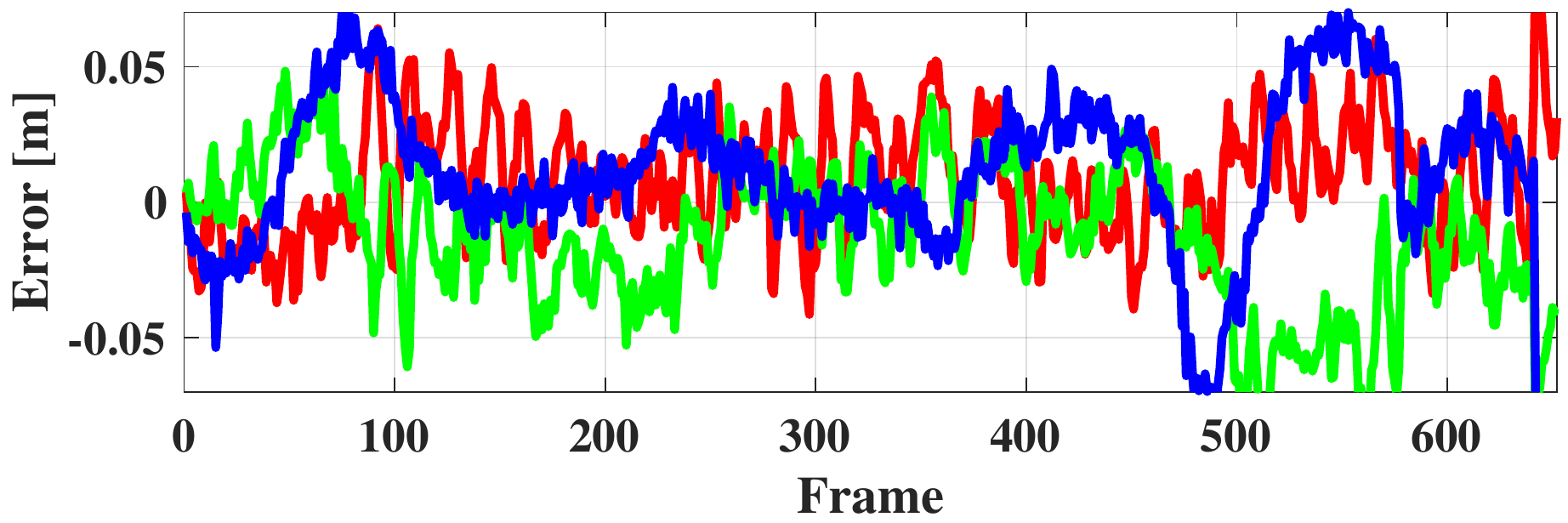}
			\end{minipage}
		}
		\subfigure[] { 
			\begin{minipage}{0.32\textwidth}
				\centering
				\includegraphics[width=1\columnwidth]{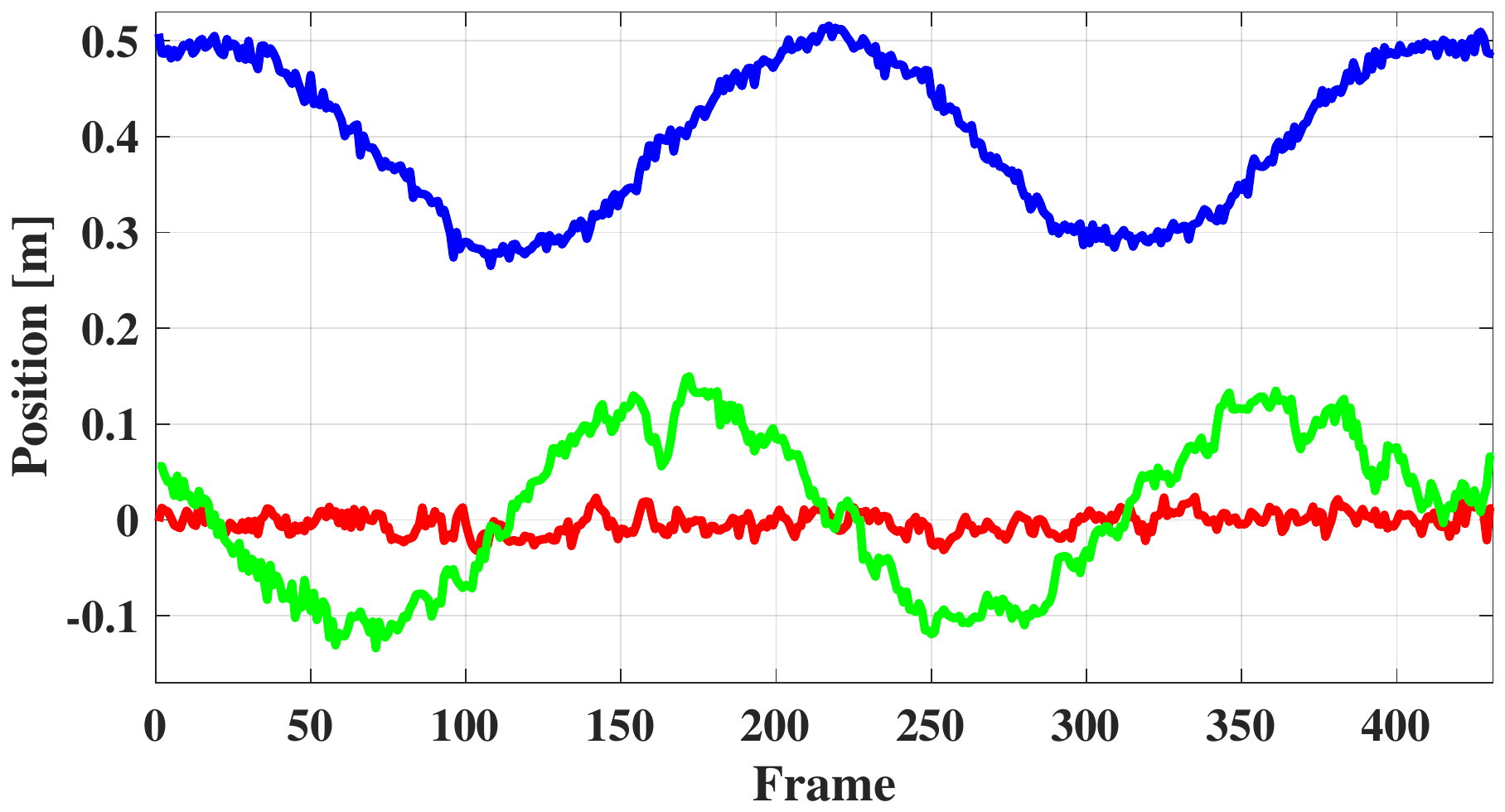}
				\\
				\includegraphics[width=1\columnwidth]{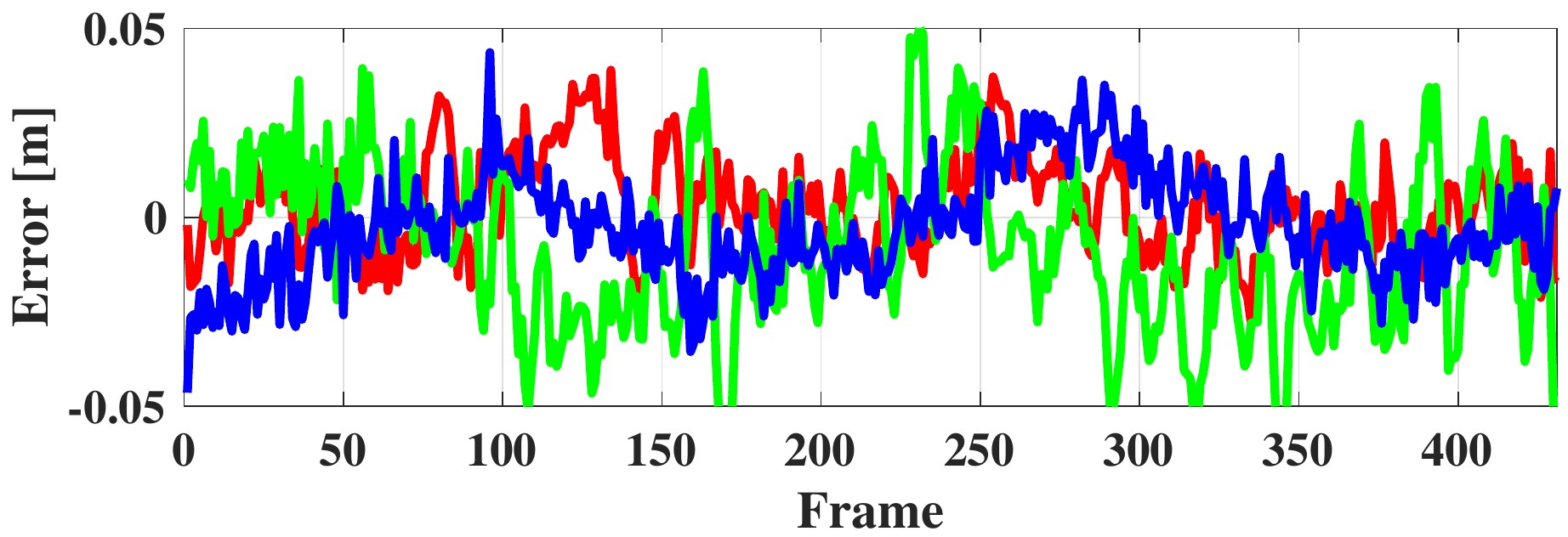}
			\end{minipage}
		}
		\subfigure[] {  
			\begin{minipage}{0.32\textwidth}
				\centering
				\includegraphics[width=1\columnwidth]{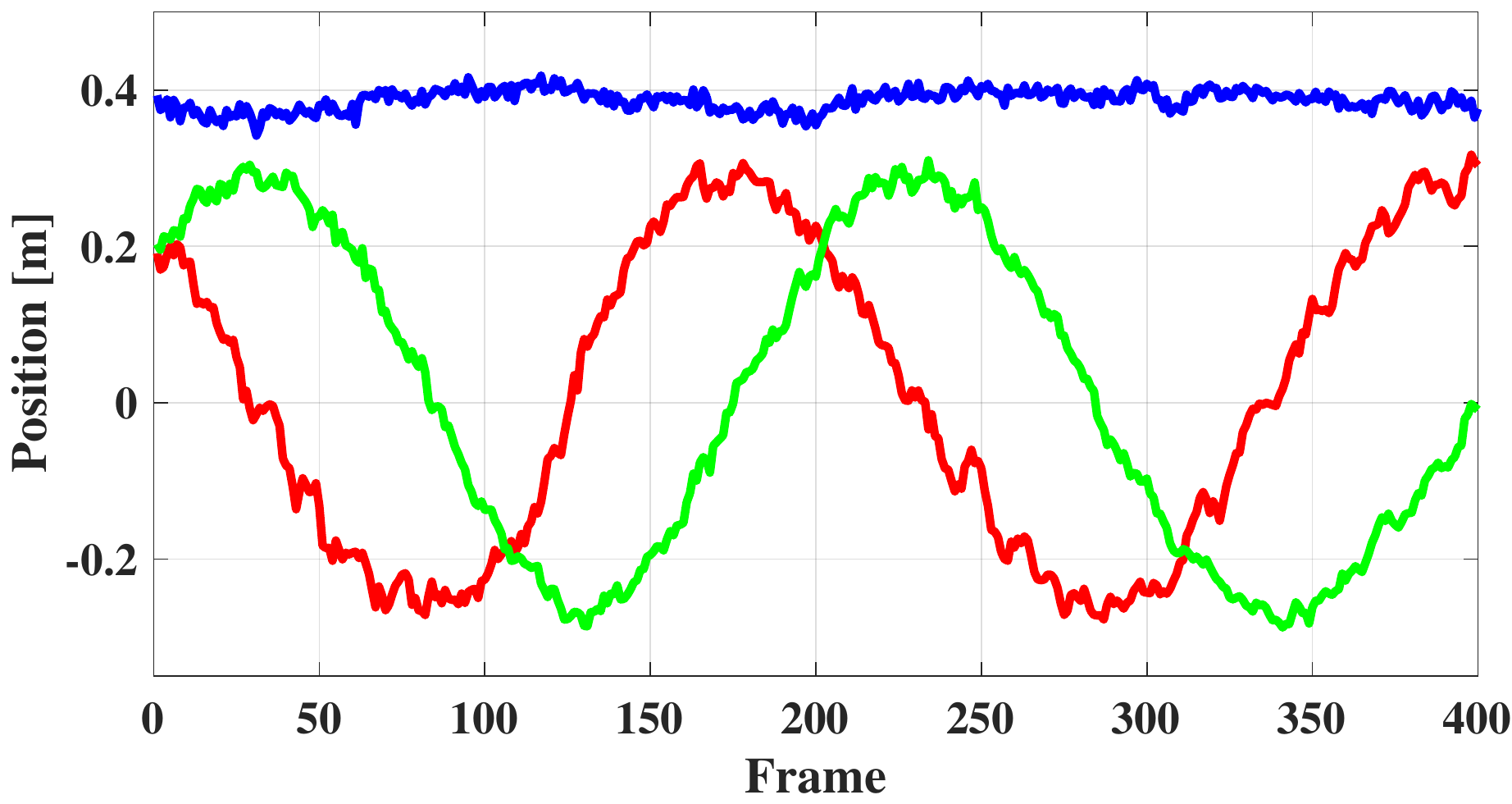}
				\\
				\includegraphics[width=1\columnwidth]{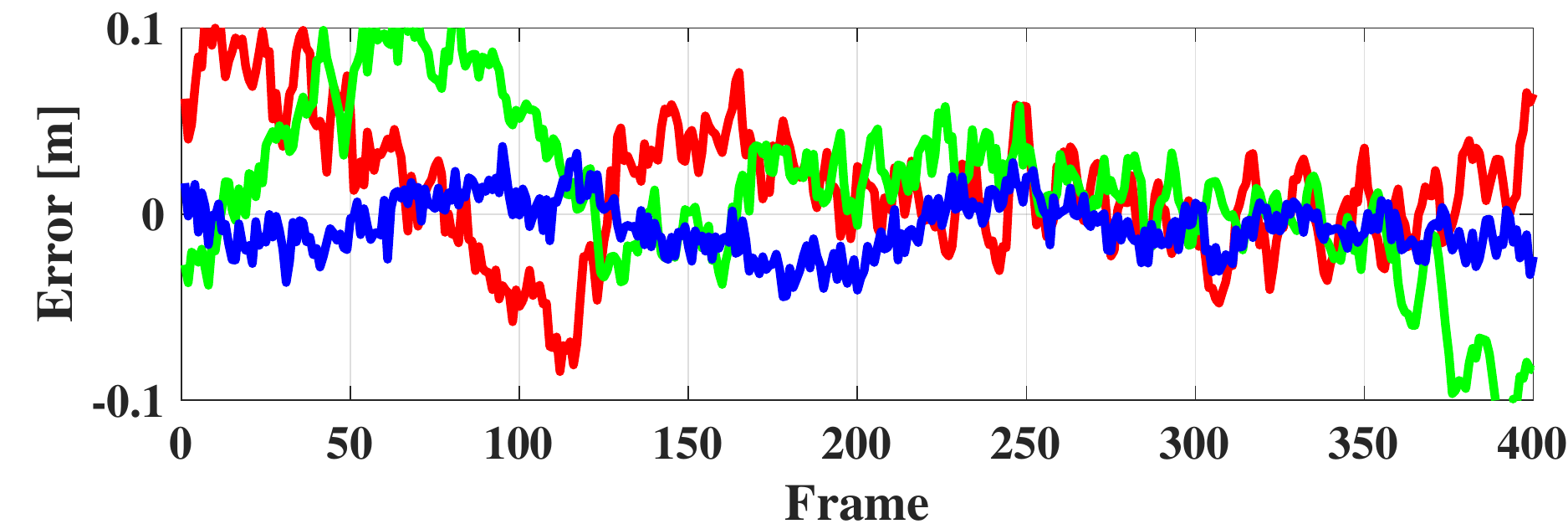}
			\end{minipage}
		} \\
		\subfigure[] { 
			\begin{minipage}{0.32\textwidth}
				\centering
				\includegraphics[width=1\columnwidth]{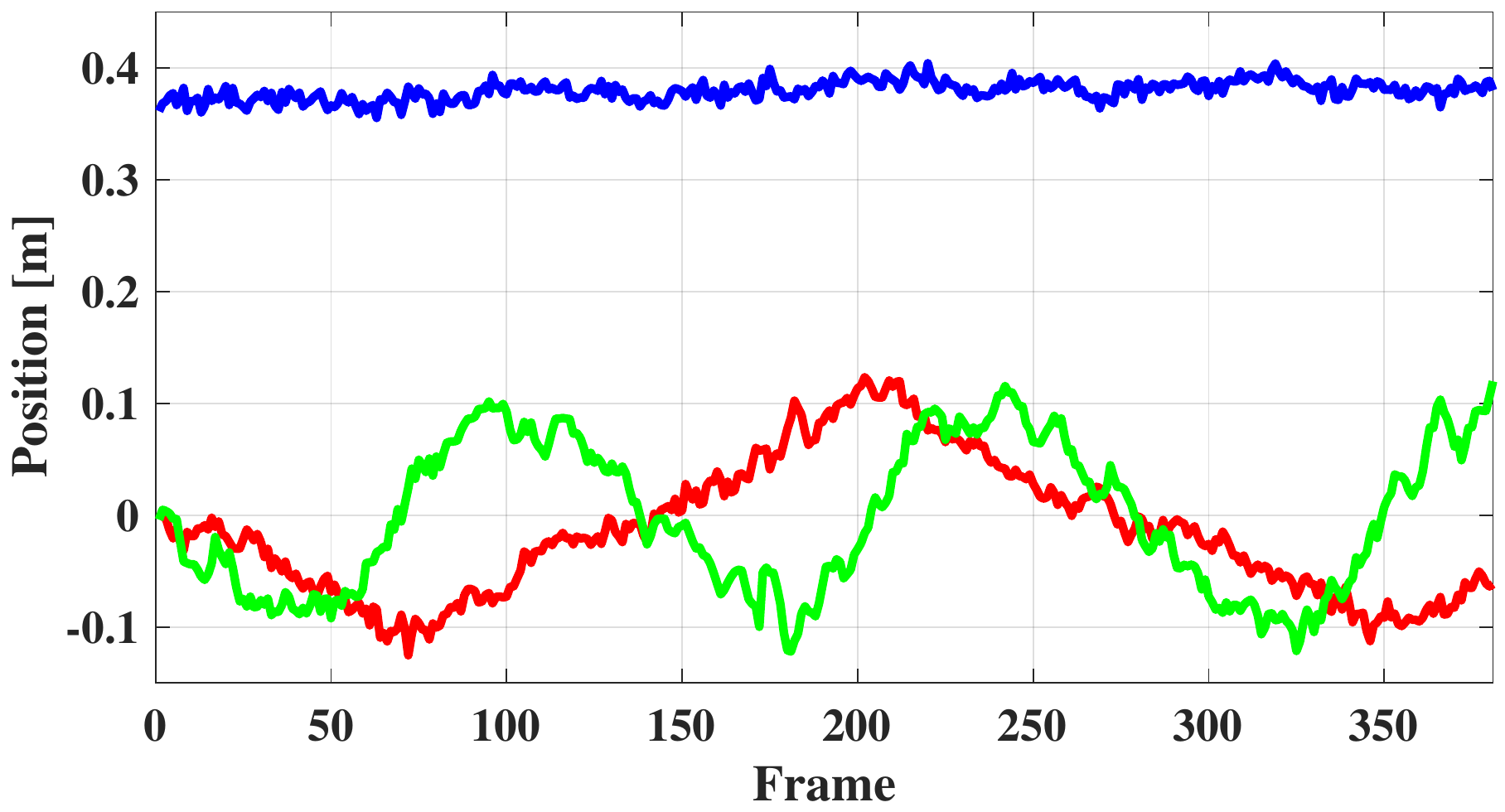}
				\\
				\includegraphics[width=1\columnwidth]{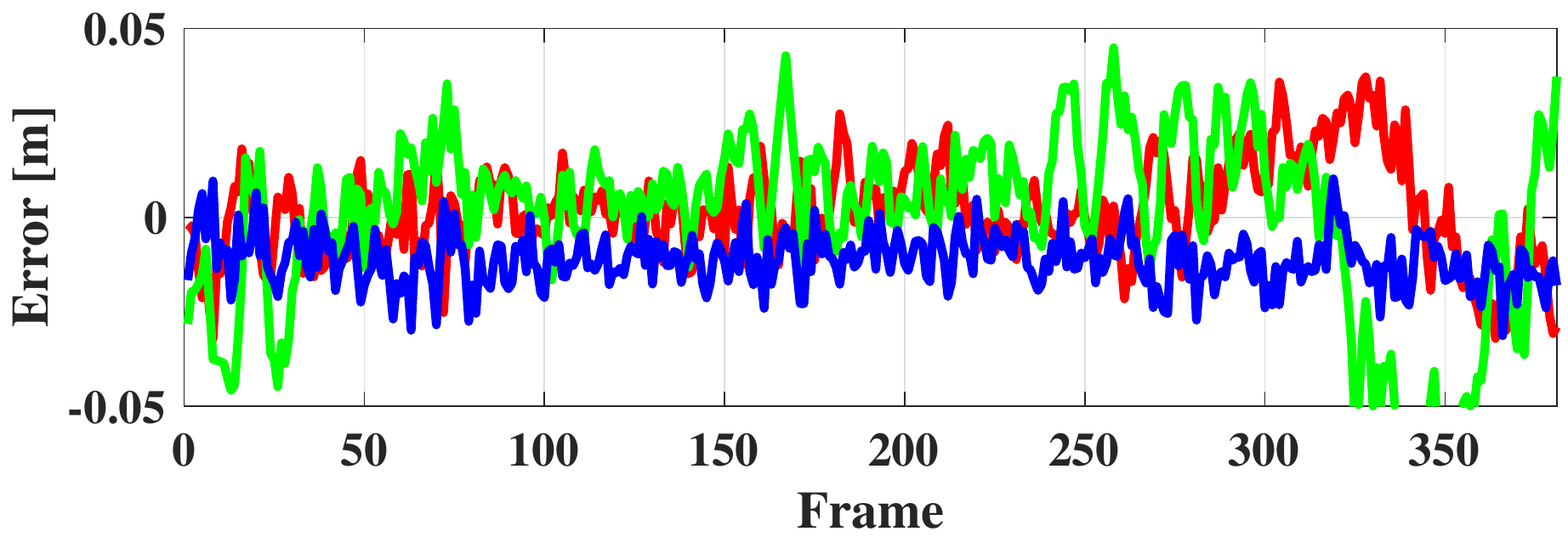}
			\end{minipage}
		}
		\subfigure[] { 
			\begin{minipage}{0.32\textwidth}
				\centering
				\includegraphics[width=1\columnwidth]{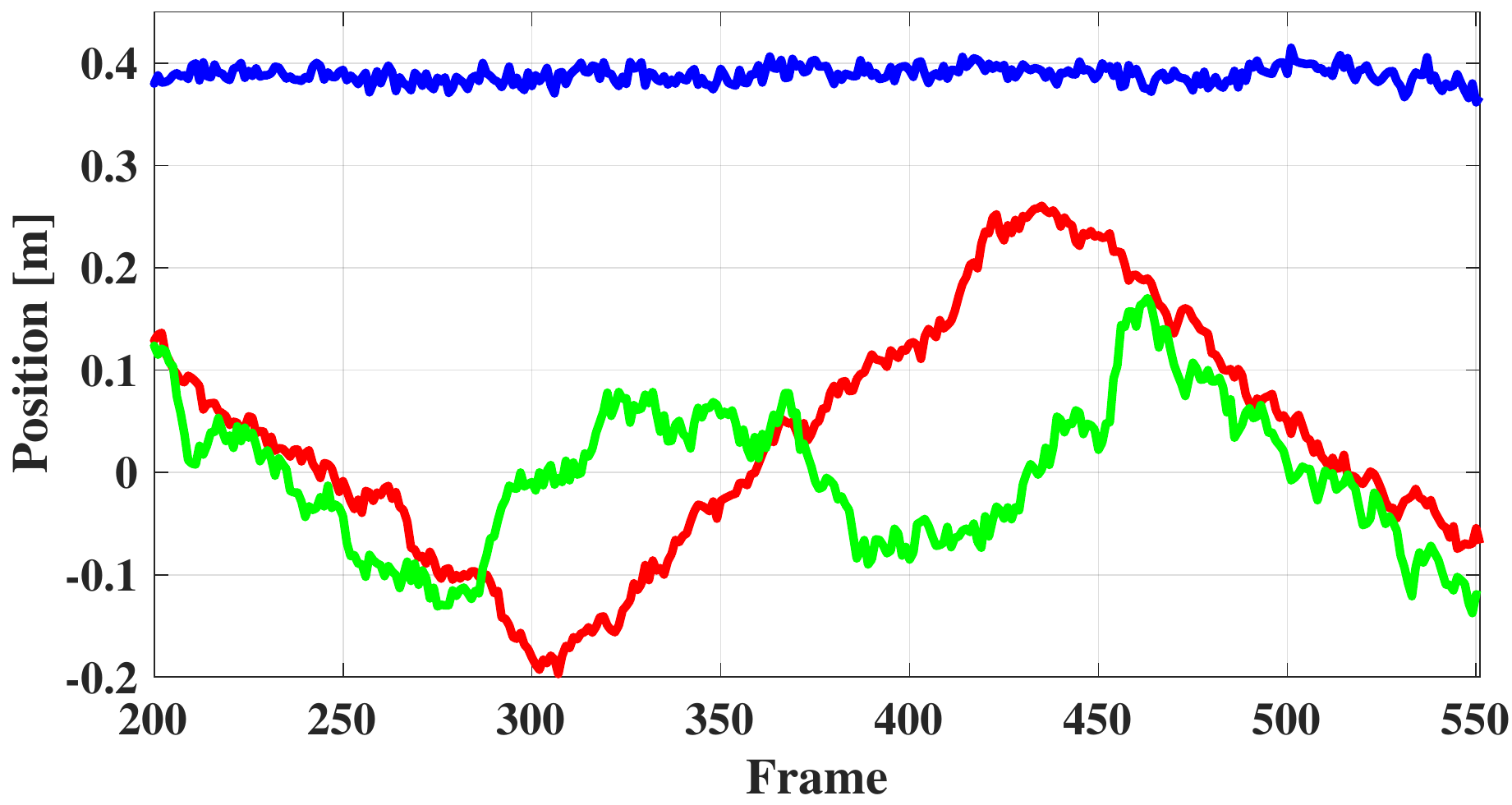}
				\\
				\includegraphics[width=1\columnwidth]{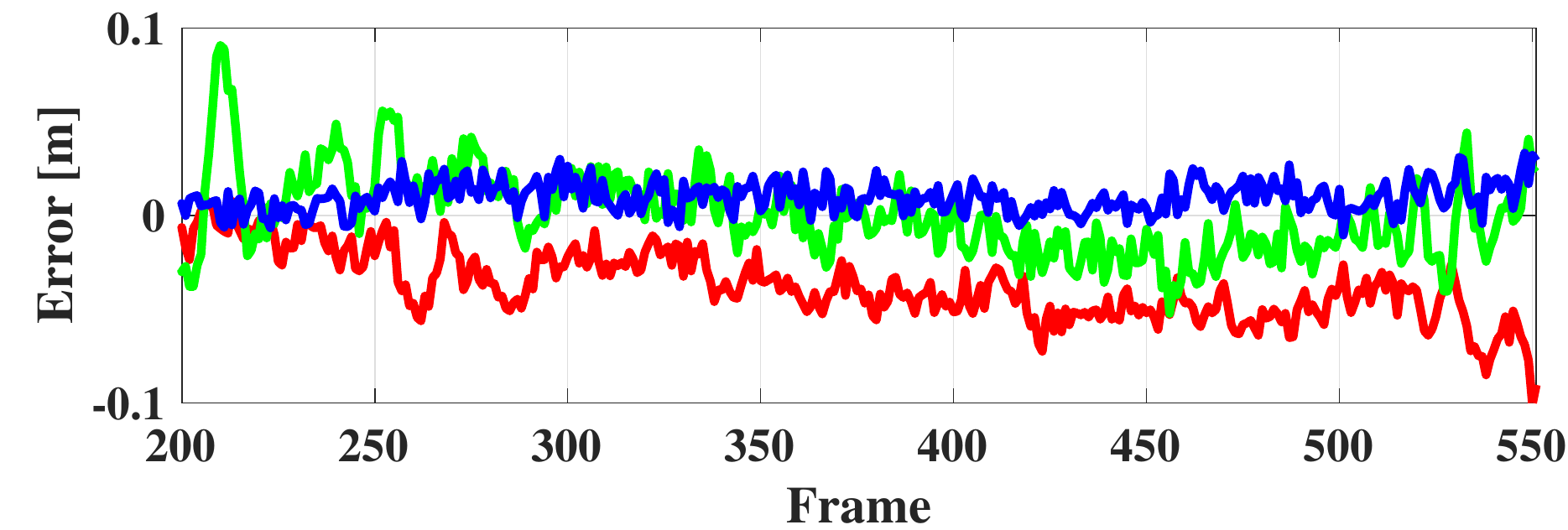}
			\end{minipage}
		}
		\subfigure[] {  
			\begin{minipage}{0.32\textwidth}
				\centering
				\includegraphics[width=1\columnwidth]{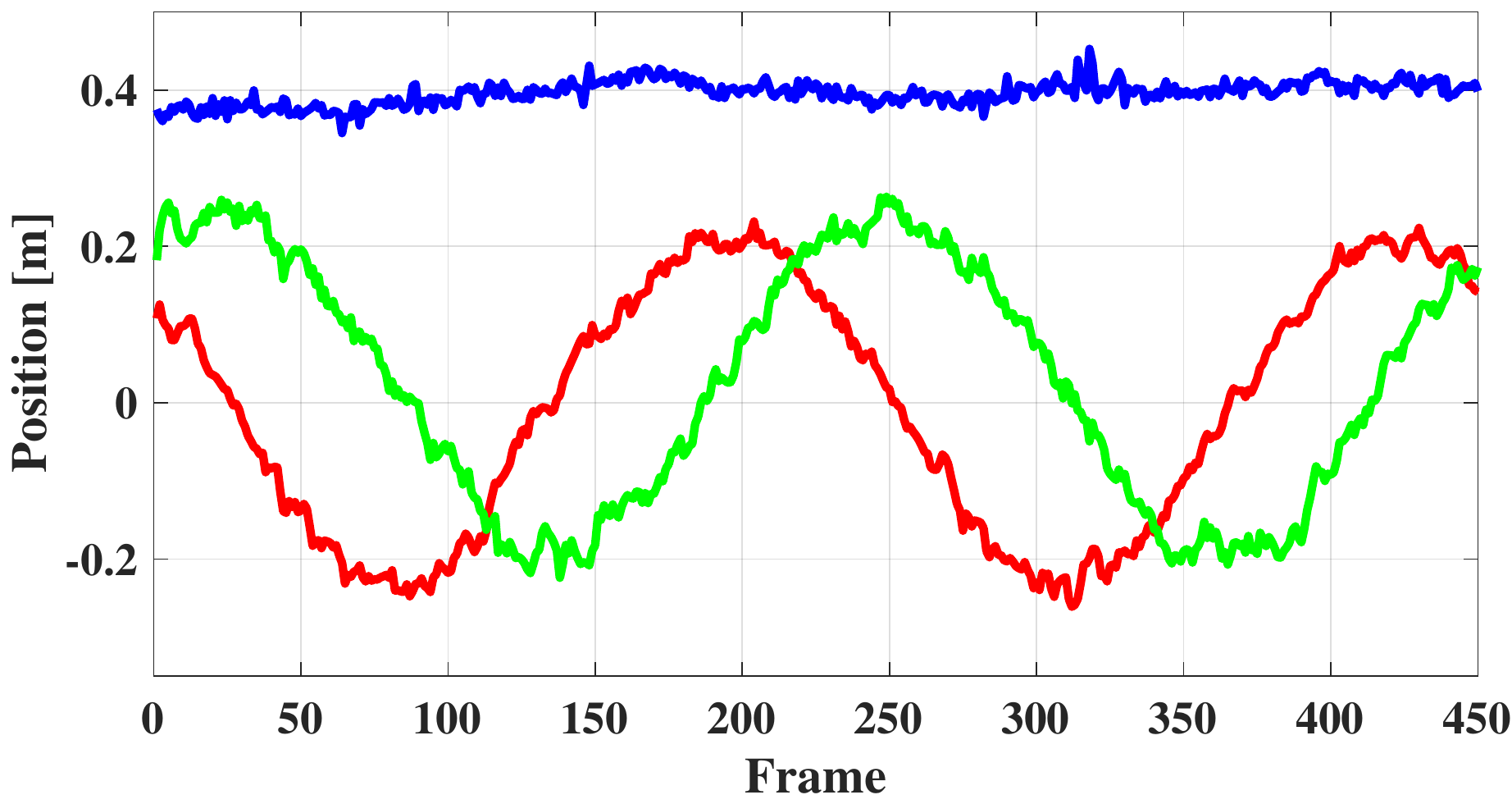}
				\\
				\includegraphics[width=1\columnwidth]{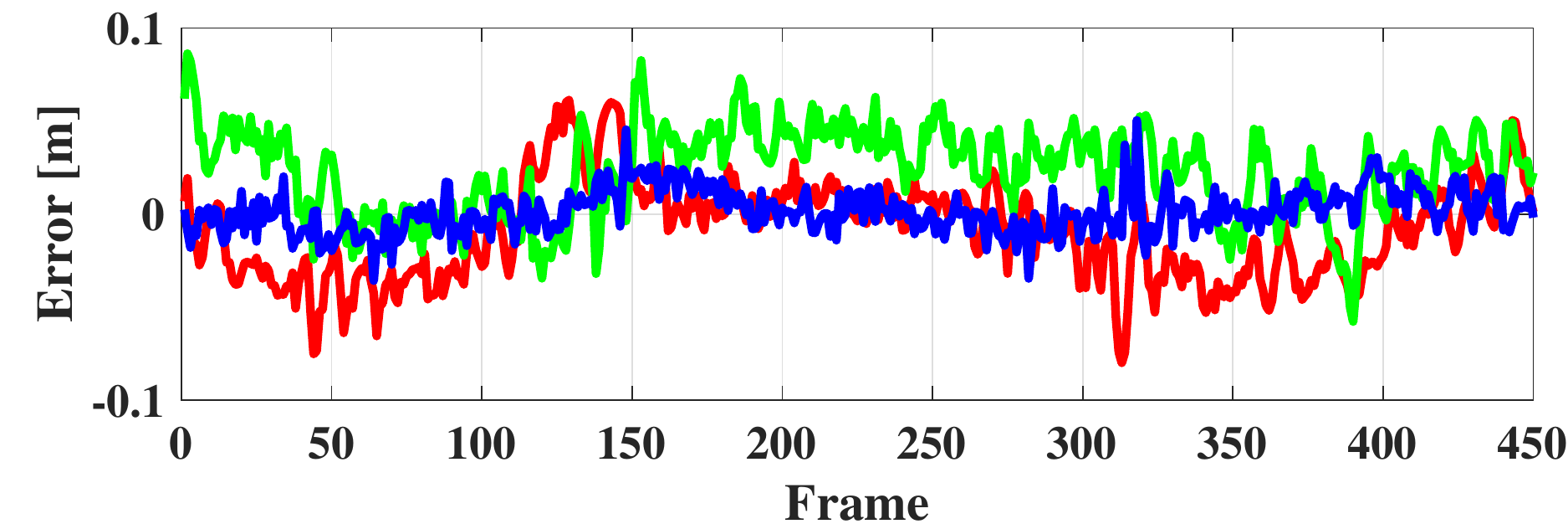}
			\end{minipage}
		}
	\end{center}
	
	\caption{Estimation of camera position and the respective errors on six datasets. Lines with \textcolor[rgb]{ 1,  0,  0}{red}, \textcolor[rgb]{ 0,  1,  0}{green} and \textcolor[rgb]{ 0,  0,  1}{blue} color denote x, y and z positions, respectively. The ground truth is not displayed because there is no noticeable differences with our results at such scale.}
	\label{fig:pose_estimation}
\end{figure*}
We evaluate our localization system on six datasets covering 2,666 images, and in each dataset, the camera is moving at a distinct trajectory as UAV flies. The image is captured with a resolution of $1280 \times 720$ pixels at 10fps, using Intel RealSense (R200) camera looking ahead to perform building inspection, as shown in Fig.~\ref{fig:localization}. 

We adopt the UAV location in motion capture system as the ground truth. The mean position errors in x, y and z directions are reported in Table~\ref{tab:pose estimation}. Figure~\ref{fig:pose_estimation} exhibits the estimated position as well as respective error in every frame. The root-mean-square error (RMSE) of our method on 2,666 frames is 3.44 centimeters. 

\noindent\textbf{Remark 4:} It is noted that our system is applicable in various scenarios because our tracker can track any arbitrary objects once given their information in the first frame. In summary, compared to LED-based localization system~\cite{Faessler2014ICRA}, our method is more versatile and can run at real-time frame rates in the real-world scenarios. 
\begin{table}[h]
	\setlength{\tabcolsep}{1.7mm}
	\small
	\centering
	\caption{Illustration of estimation errors on six datasets covering 2,666 frames. The dataset is in line with the (a)-(f) in Fig.~\ref{fig:pose_estimation}.}
	\vspace{0.3cm}
	\begin{tabular}{c c c c cc}
		\hline
		Dataset&x(cm)& y(cm)&z(cm)&RMSE&Frame number\\
		\hline	\hline
		(a)&1.90&2.25&2.38&3.79&652\\
		(b)&1.06&1.88&1.13&2.44&431\\
		(c)&3.01&3.51&1.30&4.80&400\\
		(d)&1.01&1.73&1.16&2.32&381\\
		(e)&3.77&1.77&1.05&4.30&352\\
		(f)&2.27&2.77&0.91&3.69&450\\
		
		Average&2.17&2.32&1.32&3.44&444\\
		
		\hline
	\end{tabular}%
	\label{tab:pose estimation}%
\end{table}%
\section{Conclusion}
In this work, a generally applicable automatic spatio-temporal regularization framework is proposed for high-performance UAV tracking. 
Local response variation indicates local credibility, thus restricting local correlation filter learning. Global variation is able to control how much the correlation filter learns from the whole object.
Comprehensive experiments have validated that AutoTrack is the best CPU-based tracker with a speed of $\sim$$60$fps, and even outperforms some state-of-the-art deep trackers on two UAV datasets~\cite{Li2017AAAI,Du2018ECCV}. In addition, we try to bridge the gap between the theory and practice by utilizing visual tracking in UAV localization in the real world. Considerable tests proved the effectiveness and generality of our method. We strongly believe that our work can promote the development of visual tracking and its application in robotics.

\textbf{Acknowledgment:}  This work is supported by the National Natural Science Foundation of China (No.61806148), the Fundamental Research Funds for the Central Universities (No.22120180009), and Tsinghua University Initiative Scientific Research Program.
{\small
	\bibliographystyle{unsrt}
	\bibliography{reference}
}
\end{document}